\documentclass[review,12pt]{elsarticle}

\usepackage{amssymb}
\usepackage{amsmath}
\usepackage{natbib}
\usepackage{amsfonts,amssymb} 
\usepackage{multirow}
\usepackage{makecell}
\usepackage{algorithm}
\usepackage{algpseudocode} 
\usepackage{empheq}
\usepackage[svgnames]{xcolor}
\usepackage{soul}          
\sethlcolor{yellow}        

\usepackage[colorlinks,linkcolor=red,anchorcolor=blue,citecolor=green]{hyperref}

\usepackage{longtable}

\usepackage{graphicx}  
\usepackage{float} 
\usepackage{subcaption} 
\usepackage[section]{placeins} 
\usepackage{setspace}
\usepackage{lineno}

\journal{International Journal of Greenhouse Gas Control}

\begin{document}

\begin{frontmatter}

\begin{titlepage}
\begin{center}
\vspace*{1cm}

\textbf{ \large Optimal CO$_2$ storage management considering safety constraints in multi-stakeholder multi-site CCS projects: a Markov game perspective}

\vspace{1.5cm}

Jungang Chen$^{a}$ (jungang.chen@beg.utexas.edu), \\
Seyyed A. Hosseini$^a$ (seyyed.hosseini@beg.utexas.edu) \\

\hspace{10pt}

\begin{flushleft}
\small  
$^a$ 10100 Burnet Rd., Bldg. 130 Austin, TX 78758 USA

\vspace{1cm}
\textbf{Corresponding author at: 10100 Burnet Rd., Bldg. 130, Austin, TX 78758, USA.} \\
Jungang Chen \\
10100 Burnet Rd., Bldg. 130, Austin, TX 78758, USA. \\
Tel: 512-471-0140 \\
Email: jungang.chen@beg.utexas.edu

\end{flushleft}        
\end{center}
\end{titlepage}






\title{Optimal CO$_2$ storage management considering safety constraints
in multi-stakeholder multi-site CCS projects: a Markov game perspective}

\author[beg]{Jungang Chen}

\author[beg]{Seyyed A. Hosseini}

\address[beg]{Bureau of Economic Geology, Jackson School of Geosciences,
The University of Texas at Austin, 10100 Burnet Rd., Bldg. 130,
Austin, TX 78758, USA}


\begin{abstract}
Carbon capture and storage (CCS) projects typically involve a diverse array of \textit{stakeholders} or \textit{players} from public, private, and regulatory sectors, each with different objectives and responsibilities. Given the complexity, scale, and long-term nature of CCS operations, determining whether individual stakeholders can independently maximize their interests—or whether collaborative coalition agreements are needed—remains a central question for effective CCS project planning and management. 

CCS projects are often implemented in geologically connected sites, where shared geological features such as pressure space and reservoir pore capacity can lead to competitive behavior among stakeholders. Furthermore, CO$_2$ storage sites are often located in geologically mature basins that previously served as sites for hydrocarbon extraction or wastewater disposal in order to leverage existing infrastructures, which makes unilateral optimization even more complicated and unrealistic. 

In this work, we propose a paradigm based on Markov games to quantitatively investigate how different coalition structures affect the goals of stakeholders. We frame this multi-stakeholder multi-site problem as a multi-agent reinforcement learning problem with safety constraints.  Our approach enables agents to learn optimal strategies while compliant with safety regulations. We present an example where multiple operators are injecting CO$_2$ into their respective project areas in a geologically connected basin. To address the high computational cost of repeated simulations of high fidelity models, a previously developed surrogate model based on the Embed-to-Control (E2C) framework is employed. Our results demonstrate the effectiveness of proposed framework in addressing optimal management of CO$_2$ storage when multiple stakeholders with various objectives and goals are involved. 

 
\end{abstract}




\end{frontmatter}


\section{Introduction}
\label{sec:intro}

\subsection{Background}
Geological carbon storage (GCS) has emerged as a viable technology in global efforts to mitigate climate change by permanently sequestering carbon dioxide (CO$_2$) in deep subsurface formations. As part of the broader carbon capture and storage (CCS) portfolio, GCS can help reduce anthropogenic emissions from industrial and other hard-to-abate sources and support net-zero pathways alongside rapid emissions reductions\citep{celia2015status, hosseini2024dynamic, metz2005ipcc}. Although large-scale GCS deployment remains at an early stage, projects are increasingly being implemented at the basin scale, where geological formations span large areas and support long-term operations and large-volume storage \citep{huang2014basin, birkholzer2009basin, wijaya2024basin}. These large-scale deployments often involve multiple injection sites, complex monitoring and regulatory requirements, and diverse stakeholder participation. In this context, multi-stakeholder carbon capture and storage (CCS) initiatives—where operators, landowners, regulators, and investors interact within geologically connected sites—are becoming increasingly common.

Recent basin-scale GCS studies emphasize that large-scale GCS may be limited by pressure buildup and interference among neighboring projects, motivating coordinated pressure-management strategies among operators. For example, \citet{bump2024pressure} introduced ``pressure space'' as a basin-scale limiting commodity for CCS and showed that pressure-limited resource estimates can materially reduce practical storage expectations relative to pore-volume-only calculations. \citet{jahediesfanjani2019improving} quantify pressure-limited dynamic storage capacity across multiple U.S. basins and demonstrate how brine extraction can expand pressure-limited capacity. \citet{plampin2023dynamic} provide basin-scale dynamic resource estimates highlighting regional constraints such as pressure/injectivity limits. These studies reinforce the need for decision frameworks that explicitly account for pressure constraints and multi-project interference when evaluating future basin-scale deployment scenarios.

In a multi-stakeholder CCS setting, each stakeholder pursues distinct interests and goals \citep{bentham2014managing}. For example, operators typically manage one or more injection/extraction wells with objectives such as maximizing CO$_{2}$ storage or financial returns. Landowners focus on maximizing land use value and royalty income, while regulators prioritize long-term safety, pressure containment, and environmental compliance, e.g. Class \text{VI} well regulations \citep{EPA2013UIC, leng2024comprehensive}. Although only operators physically interact with the reservoir, interests of all other stakeholders also shape the operational landscape. For instance, regulators may impose safety constraints that limit injection volumes and rates, whereas landowners and operators may both seek to exploit the available pressure space \citep{bump2024pressure} for maximum economic return. Moreover, the interconnected nature of subsurface dynamics means that one operator’s injection strategy can alter reservoir pressure and plume behavior, influencing the economic and operational outcomes of other operators.

These inter-dependencies pose significant challenges to effective basin-wide coordination and optimization. Without a robust decision-making framework that integrates individual or collective objectives with system-wide regulatory constraints, stakeholders may risk adopting locally optimal strategies that undermine collective performance, leading to underutilized reservoir capacity, reduced economic returns, or even regulatory noncompliance. To this end, coordinated multi-agent optimization is essential to balance return, safety, and fairness in basin-scale GCS deployment.


\subsection{Modeling Multi-agent Optimization in GCS}

Basin-scale GCS modeling studies increasingly emphasize that pressure buildup and inter-project interference can become the binding system constraint long before CO$_2$ plumes physically interact \citep{cunha2025basin, duggan2024managing, anderson2020estimating}, especially when multiple large injection projects operate in proximity. For example, basin-scale numerical experiments in the Illinois Basin \citep{bandilla2017active} show that active pressure management via brine production can materially reduce pressure buildup and shrink regulatory Areas of Review (AoR), highlighting how operational choices at one site can affect feasibility and monitoring requirements at others. More recent multi-well, multi-project simulations \citep{wijaya2024dynamic} explicitly track the co-evolution of pressure fronts and CO$_2$ plumes under various operational variables, demonstrating that interference can tighten allowable injection envelopes over time and motivating coordinated planning at the basin scale.

Despite extensive basin-scale modeling studies, optimal management of basin-scale geological carbon storage remains challenging, especially when multiple stakeholders with respective goals and interests are involved. On the one hand, high fidelity models that predict the pressure evolution and plume migration are needed to ensure meeting safety constraints. Optimization of this system requires multiple forward evaluations of these models, which becomes infeasible when the computational resources are limited. A remedy to this is employing machine learning-based proxy models to amortize the computation costs \citep{wang2024deep, gurwicz2025assessing}. On the other hand, the literature on multi-agent optimization for GCS or broader subsurface communities is scarce. Existing literature either only explored multi-objective optimization of one single stakeholder on various CO$_{2}$ storage projects \citep{you2020development, park2021multi, liu2025multi} or constrained co-optimization of one single stakeholder \citep{zou2023integrated, nguyen2024multi, tang2025graph}. Recently, \citep{pettersson2024multi, pettersson2025cooperative} proposed a cooperative game model with multi-objective optimization that enables coordinated CO$_{2}$ injection strategies among multiple agents in basin-scale carbon storage projects. This framework provides a general map of possible solution strategies for all possible coalition structures, but no mechanism for determining for imposing one of them rather than another. Consequently, unless there is a single centralized decision maker who assigns operational strategies to all agents, the outcome remains uncertain for all agents involved.


\subsection{Constrained Markov Game (CMG) and Safe Multi-agent Reinforcement Learning (safe MARL)}

In addition to the traditional optimization framework, which based on gradient-based or evolutionary algorithms, a framework referred to as \textbf{constrained Markov game (CMG)} \citep{altman2021constrained} is suitable for constrained optimization and offers real-time optimized solutions when multiple actors are involved. A constrained Markov game extends the Markov game framework by adding safety constraints that all agents must comply with while optimizing their own objectives. In a CMG, the system is represented as a sequential decision-making process defined by states, actions, transitions, and rewards, along with explicit safety constraints that limit feasible actions. Safe multi-agent reinforcement learning provides practical methods to learn policies in such settings, ensuring agents maximize rewards while avoiding constraint violations. The detailed formulation of this approach is presented in section \ref{sec:Prelim} and section \ref{sec:Methodo}. 

This CMG framework is particularly suited to multi-stakeholder CCS projects, where agents operate under competing objectives and must comply with safety regulations. It enables the development of policies that are not only economically optimal but also taking others' operational actions into account, thereby providing a scalable, adaptive decision-making framework for basin-scale CO$_{2}$ storage management.


\subsection{Contributions and Paper Organization}

This work develops a modeling and decision-support framework to analyze potential basin-scale,
multi-operator GCS deployment scenarios and associated pressure-management challenges. The contributions of this study are twofold. First, we develop a safe multi-agent reinforcement learning framework tailored for multi-stakeholder, multi-site CO$_{2}$ storage problems when each stakeholder has its own goals and interests. The framework accommodates these divergent interests while ensuring safe operations, and its performance is validated against traditional multi-objective optimization approaches. Second, a reward and penalty model is meticulously designed for CO$_{2}$ storage projects, and various coalition structures based on shared rewards and penalties have been investigated to evaluate their impacts on stakeholder objectives and overall system performance.

The paper is organized as follows. Section~\ref{sec:Problem} reviews the subsurface dynamics of GCS projects, including key objectives, operational constraints, and potential coalition strategies among stakeholders. Section~\ref{sec:Prelim} introduces fundamental concepts in reinforcement learning and Markov games, establishing the theoretical basis for our approach. Section~\ref{sec:Methodo} presents the proposed methodology and workflow tailored for GCS. Section~\ref{sec:Results} reports the case study results. Finally, the paper concludes with a discussion of key findings and a summary of contributions and directions for future work.

\section{Problem Formulation}
\label{sec:Problem}
\subsection{Overview of Basin-Scale GCS System}
Geological CO$_{2}$ storage typically consists of studies and analyses across various scales \citep{guo2022role, ni2021quantifying} and involves complex coupled multi-physical processes \citep{nordbotten2011geological, JIANG20113557}. Deep saline aquifers have abundant storage potential and are regarded as reliable and feasible storage sites. When CO$_{2}$ is compressed and supercritical CO$_{2}$ is being injected into deep saline aquifers, a compositional multi-phase system is formed.     In our study, we consider a compositional porous flow process where the governing equation of the system can be described as: 

\begin{equation}
\frac{\partial}{\partial t} \left( \phi (\rho_w S_w c_{i,w} + \rho_g S_g c_{i,g}) \right) 
+ \nabla \cdot (c_{i,w} \rho_w \mathbf{v}_w + c_{i,g} \rho_g \mathbf{v}_g) \\
= \sum_{j=1}^{n} \left( c_{i,w} \rho_w q_w^j + c_{i,g} \rho_g q_g^j \right)
\end{equation}

Where the first term is referred to as \textbf{accumulation term}, \( \frac{\partial}{\partial t} \) is the time differentiation, \( \phi \) stands for porosity, \( \rho_\alpha \) and \( S_\alpha \) represent phase density and saturation of phase \( \alpha = w, \, g \) respectively; the second term is referred to as \textbf{flux term}, \( c_{i,\alpha} \) denotes mole fraction of component \( i \) in phase \( \alpha \), \( \mathbf{v}_\alpha \) is the volumetric velocity of phase \( \alpha \); the third term denotes the \textbf{source/sink term}, \( q_\alpha^{j} \) denotes the volumetric rate for phase \( \alpha \) and well \( j \). \( \mathbf{v}_\alpha \) is given by Darcy’s law, which reads

\begin{equation}
\mathbf{v}_\alpha = -\mathbf{k} \frac{k_{r\alpha}}{\mu_\alpha} \left( \nabla p_\alpha + \rho_\alpha g \nabla D \right)
\end{equation}

Here \( k \) refers to the total permeability; \( k_{r\alpha}, \mu_\alpha, p_\alpha \) and \( \rho_\alpha \) are the relative permeability, viscosity, pressure, and density of \( \alpha \)-phase, respectively. \( g \) denotes the gravitational constant. 

After spatial and temporal discretizations, the above governing equations that describe the compositional flow in porous media can be reformulated in state-space form, which corresponds to the environment dynamics used in the subsequent Markov Decision Process (MDP) / Markov game formulations: 

\begin{equation}
\begin{aligned}
x_{t+1} &= f(x_t, u_t) \\
y_{t+1} &= g(x_{t+1}, u_t)
\end{aligned}
\label{eqn: statespace}
\end{equation}

where $t \in \{0,1,\dots,T-1\}$ is the discrete time index and $t+1$ denotes the next time step. $x_t \in \mathbb{R}^{2*N_b}$ is the system state (e.g., pressure and CO$_2$ composition fields), $u_t \in \mathbb{R}^{N_u}$ is the control/action (e.g., well injection/extraction rates), and $y_t \in \mathbb{R}^{N_{obs}}$ denotes the simulated outputs/observations of interest.Here, $N_b$, $N_u$ and $N_{obs}$ denote the number of grid blocks, the dimension of the control variables, and the dimension of observation data, respectively. The mappings $f(\cdot)$ and $g(\cdot)$ represent the nonlinear state transition and observation operators induced by the discretized governing equations. Because the state dimension can be very large (tens of thousands to millions), using $x_t$ directly for reinforcement learning training is computationally prohibitive.

To address the computational burden of high-fidelity reservoir simulations, we leverage a surrogate reduced-order model based on the Embed-to-Control (E2C) architecture \citep{watter2015embed, jin2020deep, coutinho2021physics, chen2024advancing, chen2025application}, which captures essential system dynamics while significantly reducing simulation time. Employing the machine learning-based reduced-order model, equation (\ref{eqn: statespace}) can be approximated as: 
\begin{equation}
\begin{aligned}
z_t &= Enc(x_t; \varphi) \\
z_{t+1} &= f_{z}(z_t, u_t; \omega_{1}) \\
\hat{x}_t &= Dec(z_t; \theta) \\
\hat{y}_{t+1} &= g_{z}(z_{t+1}, u_t; \omega_{2})
\end{aligned}
\label{eqn: reducedstatespace}
\end{equation}

where $\mathrm{Enc}(\cdot ; \varphi)$ is an encoder network that maps the high-dimensional reservoir state $x_t$ to a low-dimensional latent state $z_t \in \mathbb{R}^{n_z}$, and $\mathrm{Dec}(\cdot ; \theta)$ is a decoder network that reconstructs the state $\hat{x}_t$ from $z_t$.
The latent-space transition model $f_{z}(\cdot ; \omega_{1})$ approximates the latent dynamics $z_{t+1} = f_{z}(z_t, u_t, \omega_1)$, 
while the latent-space observation model $g_{z}(\cdot ; \omega_2)$ predicts the outputs $\hat{y}_{t+1} = g_{z}(z_{t+1}, u_t; ,\omega_2)$. The parameters $\varphi$, $\theta$, $\omega_1$, and $\omega_2$ denote the weights of the corresponding neural networks.

The E2C networks are trained offline using state--action trajectories generated from the high-fidelity reservoir simulator, with an illustrative example shown in Figure \ref{fig: e2co_plot}. Training minimizes a weighted objective of the following form:

\begin{equation}
\mathcal{L}_{\mathrm{E2C}}
= \mathcal{L}_{\mathrm{rec}}(x_t,\hat{x}_t)
+ \beta\,\mathcal{L}_{\mathrm{reg}}(z_t,z_{t+1},u_t)
+ \eta\,\mathcal{L}_{\mathrm{obs}}(y_{t+1},\hat{y}_{t+1}),
\end{equation}

where $\mathcal{L}_{\mathrm{rec}}$ is a reconstruction loss, $\mathcal{L}_{\mathrm{reg}}$ is a regularization term that promotes locally linear latent dynamics, and $\mathcal{L}_{\mathrm{obs}}$ penalizes mismatch between simulated and predicted observations. The scalars $\beta$ and $\eta$ are are weighting coefficients. Once trained, the surrogate replaces expensive simulator runs during MARL training by rolling out latent dynamics and decoding the required outputs. Prior studies \citep{chen2024advancing, chen2024optimization}have shown that this proxy model maintains high robustness while achieving desirable accuracy compared to high-fidelity simulations. 
 
\subsection{Stakeholders Objectives and Operational Conflicts}

\begin{figure}[htbp]
    \centering
    \includegraphics[width =0.8\textwidth]{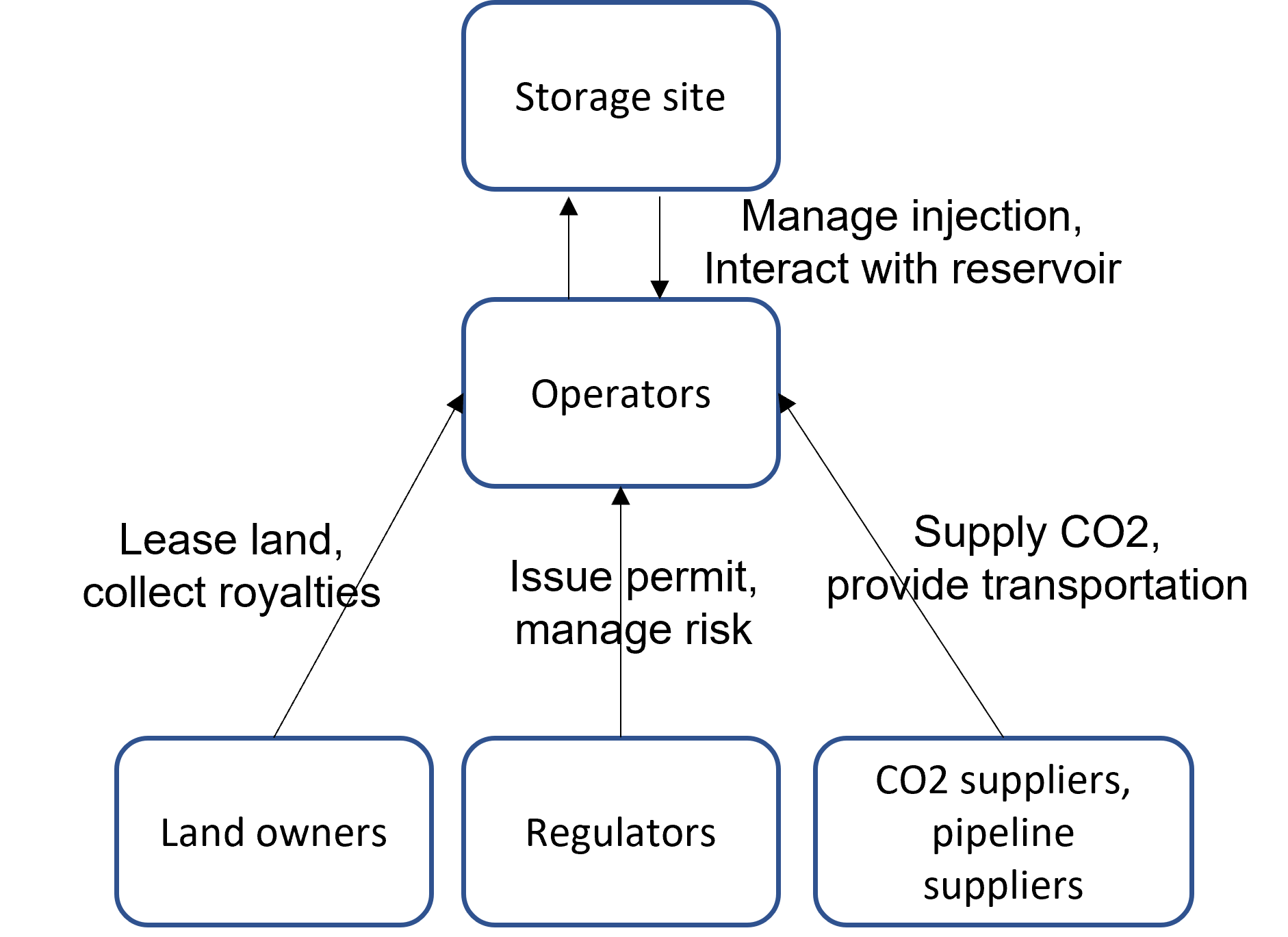}
    \caption{Relations between various stakeholders in a CCS project} 
    \label{fig: stakeholders}
\end{figure}

Basin-scale carbon capture and storage (CCS) projects involve a diverse array of stakeholders whose interests, responsibilities, and operational controls are inherently interdependent. Figure \ref{fig: stakeholders} illustrates the key actors and their roles in a typical CCS project. At the center of the system are operators, who manage CO$_{2}$ injection wells (possibly also some brine extraction wells) and directly interact with the subsurface reservoir. Their primary objectives typically include maximizing CO$_{2}$ storage volumes and financial returns through strategic control of injection rates and well operations.

Surrounding the operators are other critical stakeholders who, although not directly interacting with the reservoir, exert significant influence over operational decisions. Landowners lease the surface or subsurface rights necessary for storage operations and seek to maximize land-use value and royalty income. Regulators serve a regulatory role, issuing permits and enforcing constraints to ensure long-term safety, pressure containment, and environmental compliance. In addition, CO$_{2}$ suppliers and pipeline operators play a logistical role by delivering captured CO$_{2}$ and ensuring transportation infrastructure reliability. 

These stakeholders operate within a shared geologic basin and sometimes have misaligned goals. For instance, while both operators and landowners may favor maximizing pressure space for economic benefit, regulators may impose strict injection limits to prevent caprock fracturing or fluid migration beyond permitted boundaries. Moreover, one operator’s injection strategy can affect reservoir pressure distributions and CO$_{2}$ plume migration patterns, potentially undermining neighboring operators’ performance or violating shared safety constraints.

\subsection{Mathematical Formulation of Objectives}

In this study, we consider the primary objective to be the maximization of financial returns. Each agent’s financial outcome is quantified using present value (PV). Accordingly, for operator $i$, the present value—or immediate reward—can be expressed as:: 
\begin{equation}
\begin{aligned}
\text{PV}_t^i(u) =  \Bigg(
    & \sum_{j=1}^{N_{\text{inj}}^i} (R_{\text{credit}} - R_{\text{op}})* q_{t,j}^i(u) \\
    & + \sum_{j=1}^{N_{\text{prod}}^i} -R_{\text{w}} * q_{t,w}^i(u)
    + \sum_{j=1}^{N_{\text{prod}}^i} -R_{\text{CO}_2}* q_{t,g}^i(u)
\Bigg) \cdot \Delta t
\end{aligned}
\label{eqn: reward}
\end{equation}

where the first term represents the revenue from CO$_{2}$ credit subtracted by operational costs, the second term indicates the cost to handling the brine water, while the third term is the cost when CO$_{2}$ is re-extracted from extraction wells. The parameters $R_{\text{credit}}$, $R_{\text{op}}$, $R_{\text{w}}$ and $R_{\text{CO}_2}$ are affected by a variety of economic, policy and technological factors, such as interest rates, carbon incentive policies, brine water handling cost, and etc. To simplify the problem, we assume these parameters remain constant, as shown in table \ref{tab:revenue&cost}. The net present value (NPV) of each agent {i} is the total cumulative value discounted to current time, which can be expressed as 

\begin{equation}
\begin{aligned}
\text{NPV}^i  = \sum_{t=0}^{T} \gamma_t \cdot \text{PV}_t^i
\end{aligned}
\end{equation}

Furthermore, to avoid caprock fracturing and geomechanical risks, each company needs to satisfy the safety constraints, which can be expressed as: 
    
\begin{equation}
\begin{aligned}
\text p^i_{grid} \leq \alpha^i *p^i_{frac}
\end{aligned}
\label{eqn: eqn1}
\end{equation}

where $p_{grid, i}$ indicates the pressure at all grid blocks in agent $i$'s domain, while $p_{frac, i}$ indicates the fracturing pressure for agent $i$. Note that $p_{grid, i}(u)$ is a function of joint actions by all companies. 

\subsubsection{Multi-objective optimization formulation}
Intuitively, optimal CO$_2$ storage under multi-stakeholder settings can be formulated as a constrained multi-objective optimization (CMOO) problem, where the objective is to maximize all companies' NPVs while adhering to the pressure constraints. The constrained multi-objective optimization problem can be formulated as: 
\begin{equation}
\begin{aligned}
    \max_{u} \quad & \left\{ \text{NPV}^1(u),\, \text{NPV}^2( u),\, \ldots,\, \text{NPV}^n(u) \right\} \\
    \text{s.t.} \quad & u_{\min} \leq u \leq u_{\max}, \\
                      & p_{grid}^i(u) \leq \alpha *p_{frac}^i, i=1,2,...,n\\
                      & x_{t+1} = f(x_t, u_t), \\
                      & y_{t+1} = g(x_{t+1}, u_t), \\              
\end{aligned}
\end{equation}

where $NPV^{i}$ indicates the NPV of agent $i$, and the decisions $u$ are bounded by operational limits, pressure constraints, and the governing system transition dynamics.

\begin{figure}[htbp]
    \centering
    \includegraphics[width =0.8\textwidth]{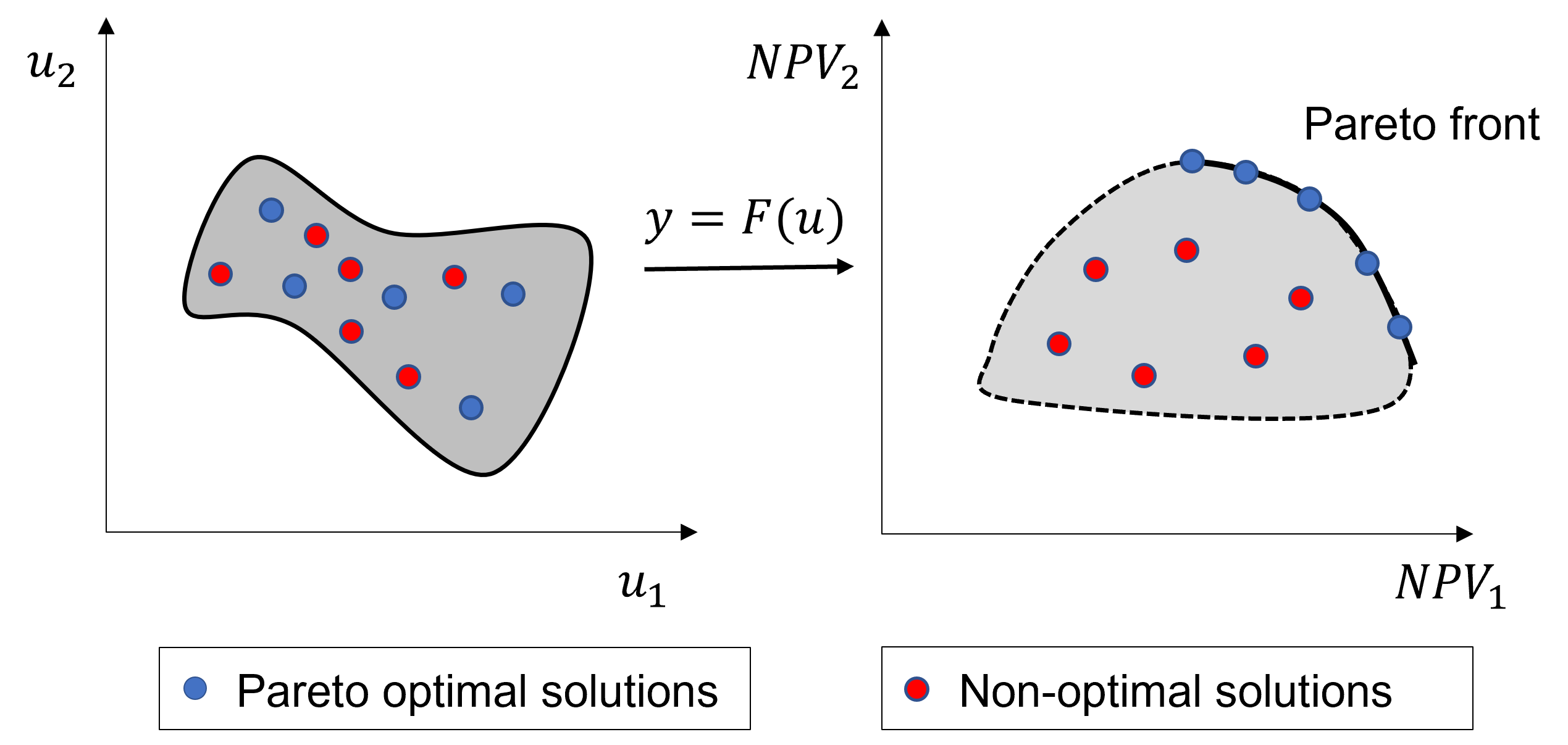}
    \caption{Sketch plot of MOO with two decision variable and two objectives. The gray areas represents the feasible decision and solution space. The blue dots in the objective space are Pareto optimal solutions forming the Pareto front, where improving one objective (an agent’s reward) necessarily degrades the other. Adapted from \citep{pereira2022review}} 
    \label{fig: moo_pareto}
\end{figure}

Traditionally, CMOO problems are addressed using evolutionary algorithms to obtain a Pareto front (see Figure \ref{fig: moo_pareto}), which provides a spectrum of optimal trade-offs for decision-making. One widely used method is the Non-dominated Sorting Genetic Algorithm II (NSGA-II) \citep{deb2013evolutionary}, implemented using the PyMOO framework in this work \citep{pymoo}. In this approach, the algorithm iteratively evolves a population of solutions to converge toward the Pareto-optimal set. Each point on the Pareto front represents a trade-off where no objective can be improved without degrading another. This front can be used for post-optimization solution selection, typically choosing a “knee point” or preference-driven solution.

Despite its effectiveness, the conventional CMOO framework may not fully capture the operational realities of CCS projects, as it assumes a single centralized decision maker with complete knowledge of basin-scale reservoir conditions, inter-well interactions, and the operational strategies of all stakeholders. In practice, CCS operations often involve multiple independent operators, each with limited information and potentially competing interests, making such centralized control unrealistic. Furthermore, CMOO typically produces a set of non-dominated solutions (a Pareto front). A separate decision rule is generally required to select a single implementable operating strategy.In multi-operator CCS settings, this introduces three practical limitations: (1). constructing a well-resolved Pareto front can be numerically expensive (and in some cases prohibitively so) because it requires many forward model evaluations. (2). the Pareto front enumerates all possible solutions under centralized optimization, but it does not specify what outcome will actually occur under decentralized decision making. (3). commonly used point-selection approaches (e.g., knee-point selection or preference-weighted scalarization) implicitly encode a choice criterion that may advantage particular stakeholders, and may not align with what is operationally feasible given contractual, regulatory, or organizational constraints  
These limitations reduce the representativeness of conventional CMOO for modeling real
multi-operator CCS decision-making environments.

\subsubsection{Constrained Markov game formulation}
To address the inherent interdependencies and competing objectives in multi-stakeholder geological carbon storage (GCS), we reformulate the optimization problem as a constrained Markov game as illustrated in Figure \ref{fig: cmarl}. Specifically, we apply multi-agent reinforcement learning (MARL) to learn an optimal policy for each stakeholder operating in the shared reservoir environment. In this framework, each operator is represented as an autonomous agent that interacts with the shared reservoir environment while pursuing its individual economic goals subject to safety constraints. 
The goal of each agent is to maximize its long-term return, defined as:
\begin{equation}
\label{formula: marl_obj}
\begin{aligned}
    \max_{\pi^{i}} \; J_{i}\!\left(\pi^{i}, \pi^{-i}\right)
    := \mathbb{E}\!\left[
    \sum_{t=0}^{T} \gamma_{t} r_{t}^{i}
    \;\middle|\;
    s_{0},\; a_{t}^{i} \sim \pi^{i}(\cdot \mid s_{t}),\; a_{t}^{-i} \sim \pi^{-i}(\cdot \mid s_{t})
    \right] 
\end{aligned}
\end{equation}
where $\pi^{i}$ denotes the policy of agent $i$, $\pi^{-i}$ represents other agents' policies except agent $i$, $\gamma_{t}$ is the discount factor, $r_t^{i}$ is the immediate reward (PV) received by agent $i$ through interacting with the reservoir environment, and $a_t^{i}$ represents the actions from operator/agent i, mapped from the respective policies conditioned on the current state $s_t$. More details of this formulation have been presented in section \ref{subsec: marl}.

In constrained Markov game, each agent $i$ is associated with a cost/penalty function $c^i_t$, representing violations of 
operational safety thresholds (e.g., pore pressure exceeding allowable threshold pressure). Analogous to the reward $r_t^{i}$, this penalty $c^i_t$ depends on current state $s_{t}$, action $a_{t}$ and subsequent state $s_{t+1}$. Collectively, the objective of this constrained Markov game formulation thus becomes:
\begin{equation}
\label{formula: cmarl_obj}
\begin{aligned}
    \max_{\pi^{i}} \quad &  J_{i}\!\left(\pi^{i}, \pi^{-i}\right) \\
    \text{s.t.} \quad & \mathbb{E}\!\left[\sum_{t=0}^{T} \gamma^{t} c_{t}^{i}    \;\middle|\; s_{0},\; a_{t}^{i} \sim \pi^{i}(\cdot \mid s_{t}),\; a_{t}^{-i} \sim \pi^{-i}(\cdot \mid s_{t}) \right]\leq d^i, \\              
\end{aligned}
\end{equation}
where $d^i$ denotes the maximum allowable cumulative violation for agent $i$. 

The constrained Markov game paradigm enables real-time, closed-loop decision making in basin-scale carbon storage, where storage operations are sequential, state-dependent. Operators' actions must adapt to evolving subsurface conditions (e.g., pressure and plume migration) and remain feasible under safety limits. The constrained Markov game directly models inter-operator interactions and learns decentralized policies that map observations/states to actions while enforcing regulatory or safety constraints.

\section{Background on Markov Games and Multi-Agent Reinforcement Learning}
\label{sec:Prelim}
\subsection{Deep Reinforcement Learning (DRL)}
Deep Reinforcement Learning (DRL) is a branch of machine learning designed for sequential decision-making problems where labeled training data are unavailable and agents must learn optimal behaviors through repeated interactions with the environment. DRL has registered tremendous success in solving various sequential decision-making problems, including but not limited to: robotics control and maneuver \citep{kober2013reinforcement,li2024deep, yu2025machine}, mechanical system design \citep{chen2025cost}, large language models \citep{havrilla2024teaching} and so on. A DRL framework is typically formalized as a Markov Decision Process (MDP) (Figure \ref{fig:rl and marl} (a)), in which an agent observes the current state $s$, selects an action $a$, and receives a corresponding reward $r$. The objective is to learn a policy that maximizes the cumulative expected reward over a defined time horizon.

A standard \textbf{Markov Decision Process (MDP)} is defined by the tuple

\[
\langle \mathcal{S}, \mathcal{A}, \mathbb{T}, r, \gamma \rangle,
\]

where:
\begin{itemize}
    \item $\mathcal{S}$ is the set of states,
    \item $\mathcal{A}$ is the set of actions,
    \item $\mathbb{T}(s' \mid s, a)$ is the transition function to state $s'$ given state $s$ and action $a$, note that for stochastic processes, it's represented by a transition probability function $\mathbb{P}(s' \mid s, a)$, 
    \item $r(s, a)$ is the immediate reward,
    \item $\gamma \in [0, 1)$ is the discount factor.
\end{itemize}

At each time step $t$, the agent observes $s_t \in \mathcal{S}$, selects an action $a_t \in \mathcal{A}$ 
based on a policy $\pi(a \mid s)$, and receives reward $r_t$ from the environment. 
The goal is to maximize the expected cumulative discounted return over the defined time horizon T:

\begin{empheq}{equation}
    \mathbb{E} \left[ \sum_{t=0}^{T} \gamma^t r_t 
    \;\Big|\; a_t \sim \pi(\cdot \mid s_t),\, s_0 \right]
\end{empheq}

Accordingly, it is convenient to summarize the long-term consequence of decisions through value functions.
The \emph{state-value} function under policy $\pi$ is defined as the expected discounted return starting
from state $s$,
\begin{equation}
V_{\pi}(s) := \mathbb{E}\!\left[\sum_{t=0}^{T}\gamma^{t} r_t \,\middle|\, s_0=s,\; a_t \sim \pi(\cdot \mid s_t)\right],
\end{equation}
and the \emph{action-value} function is the expected discounted return starting from $(s,a)$,
\begin{equation}
Q_{\pi}(s,a) := \mathbb{E}\!\left[\sum_{t=0}^{T}\gamma^{t} r_t \,\middle|\, s_0=s,\; a_0=a,\; a_t \sim \pi(\cdot \mid s_t)\right].
\end{equation}
Intuitively, $V_{\pi}(s)$ evaluates how good it is to be in state $s$ under $\pi$, while $Q_{\pi}(s,a)$
evaluates how good it is to take action $a$ at state $s$ and then follow $\pi$, which directly supports action selection and policy improvement. In this context, value-based methods learn $V_{\pi}$ or $Q_{\pi}$ from data and derive a policy by selecting actions that maximize the estimated value (e.g., $\arg\max_a Q(s,a)$), whereas policy-based methods parameterize the policy through $\pi_{\theta}$ and optimize the policy parameters $\theta$ directly (typically via policy gradients) without requiring an explicit maximization over actions; many modern actor--critic algorithms combine both by using a learned value function to guide and stabilize policy updates.

\subsection{Multi-Agent Reinforcement Learning (MARL)}
\label{subsec: marl}
\begin{figure}[htbp]
    \centering
    \begin{minipage}[t]{0.45\textwidth}
        \centering
        \includegraphics[width=1.0\textwidth]{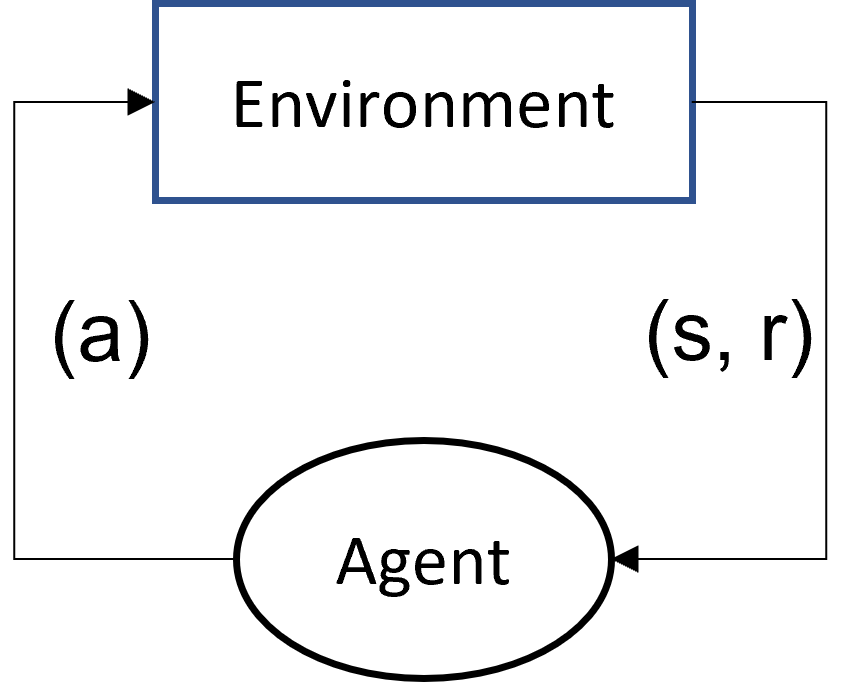}
        \vspace{0.5em}
        \small (a) Markov decision process
    \end{minipage}
    \hfill
    \begin{minipage}[t]{0.45\textwidth}
        \centering
        \includegraphics[width=1.0\textwidth]{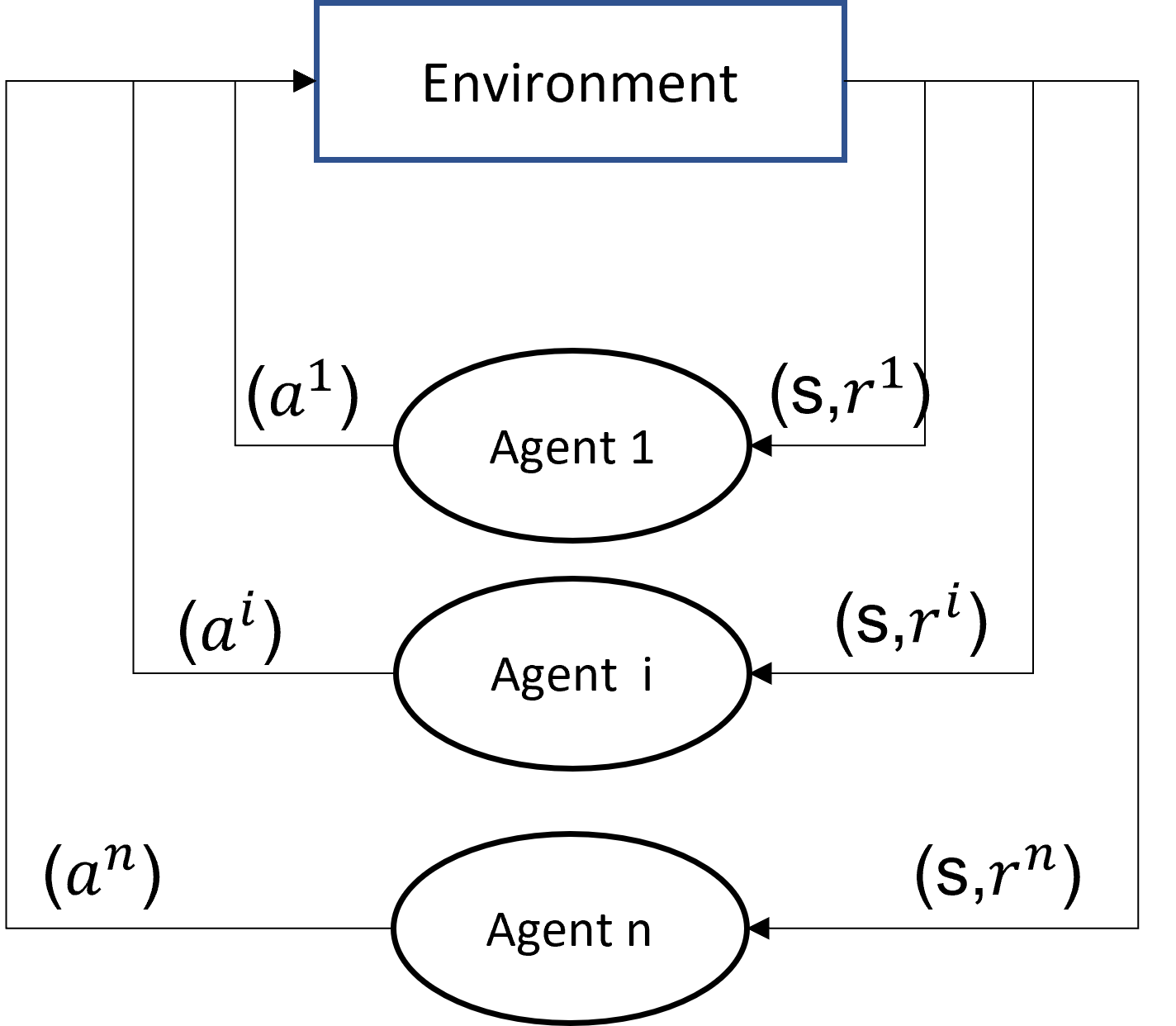}
        \vspace{0.5em}
        \small (b) Markov game
    \end{minipage}

    \caption{Schematic diagrams of (a) Markov decision process and (b) Markov game,  which correspond to the frameworks for (a) single- and (b) multi-agent RL, respectively. Adapted from \citep{zhang2021multi}}
    \label{fig:rl and marl}
\end{figure}

Traditional DRL algorithms cannot cope with multiple agents as the environment becomes non-stationary. When multiple agents interact within the same environment, it becomes a Multi-Agent Reinforcement Learning (MARL) problem \citep{tampuu2017multiagent, canese2021multi}, which is often formulated as a Markov game as in Figure \ref{fig:rl and marl} (b) \citep{wang2022cooperative}. In MARL, each agent receives observations $o_i$, takes actions $a_i$, and obtains rewards $r_i$, while the environment evolves according to the joint actions $[a_1, ..., a_i, ...a_n]$ of all agents. 

In a \textbf{Markov game} setting with $n$ agents, the framework is defined as

\[
\langle n, \mathcal{S}, \{\mathcal{A}^i\}_{i=1}^n, \mathbb{T}, \{r^i\}_{i=1}^n, \gamma \rangle,
\]

where each agent $i \in \{1, \ldots, n\}$ has:
\begin{itemize}
    \item an action $a^{i}$ from action space $\mathcal{A}^i$,
    \item a reward function $r^i(s, a^1, \ldots, a^n)$,
    \item a policy $\pi^{i}(a^i \mid s)$.
\end{itemize}

The environment transitions according to the joint action 
$a = (a^1, \ldots, a^n)$:

\begin{equation}
s_{t+1} \sim \mathbb{T}(\cdot \mid s_t, a_t^1, \ldots, a_t^n).
\end{equation}

In a multi-stakeholder CCS setting, the transition function should comply with subsurface dynamics and can be approximated by a surrogate model as in equation \ref{eqn: reducedstatespace}.
The objective of each agent $i$ in MARL framework is to maximize its own long-term return by finding a state-conditioned policy $\pi^i(\cdot \mid s)$ while considering joint policy $\pi^{-i}(\cdot \mid s)$ of all other agents except $i$, as presented in equation \ref{formula: marl_obj}. 

Similarly, the agent-specific state-value functions and action-value functions can be defined as follows: 

\begin{equation}
V^{i}(s) := \mathbb{E}\!\left[
\sum_{t=0}^{T}\gamma^{t} r_{t}^{i}
\;\middle|\;
s_{0}=s,\; a_{t}^{i}\sim \pi^{i}(\cdot\mid s_{t}),\; a_{t}^{-i}\sim \pi^{-i}(\cdot\mid s_{t})
\right],
\end{equation}

\begin{equation}
Q_r^{i}(s,a^{i},a^{-i}) := \mathbb{E}\!\left[
\sum_{t=0}^{T}\gamma^{t} r_{t}^{i}
\;\middle|\;
s_{0}=s,\; a_{0}^{i}=a^{i},\; a_{0}^{-i}=a^{-i},\; a_{t}^{i}\sim \pi^{i}(\cdot\mid s_{t}),\; a_{t}^{-i}\sim \pi^{-i}(\cdot\mid s_{t})
\right].
\end{equation}

Agents may cooperate, compete, or operate under mixed motives. Cooperation aims to maximize a shared reward, while competitive agents maximize their own objectives. In general, such interactions are general-sum, where there is no restrictions imposed on the goals of agents and their rewards can be arbitrarily related \citep{hu2003nash, singh2000nash}. The competitive CCS scenario considered in this work is in the general-sum sense, since each operator maximizes its own NPV and the sum of total NPV is not fixed. This structure makes MARL well-suited for modeling multi-stakeholder carbon storage projects, where operators, landowners, and regulators pursue distinct but interconnected goals. 

\subsection{Multi-Agent Reinforcement Learning (MARL) with Safety Constraints}
\label{subsec: cmarl}

\begin{figure}[htbp]
    \centering
    \includegraphics[width=0.5\textwidth]{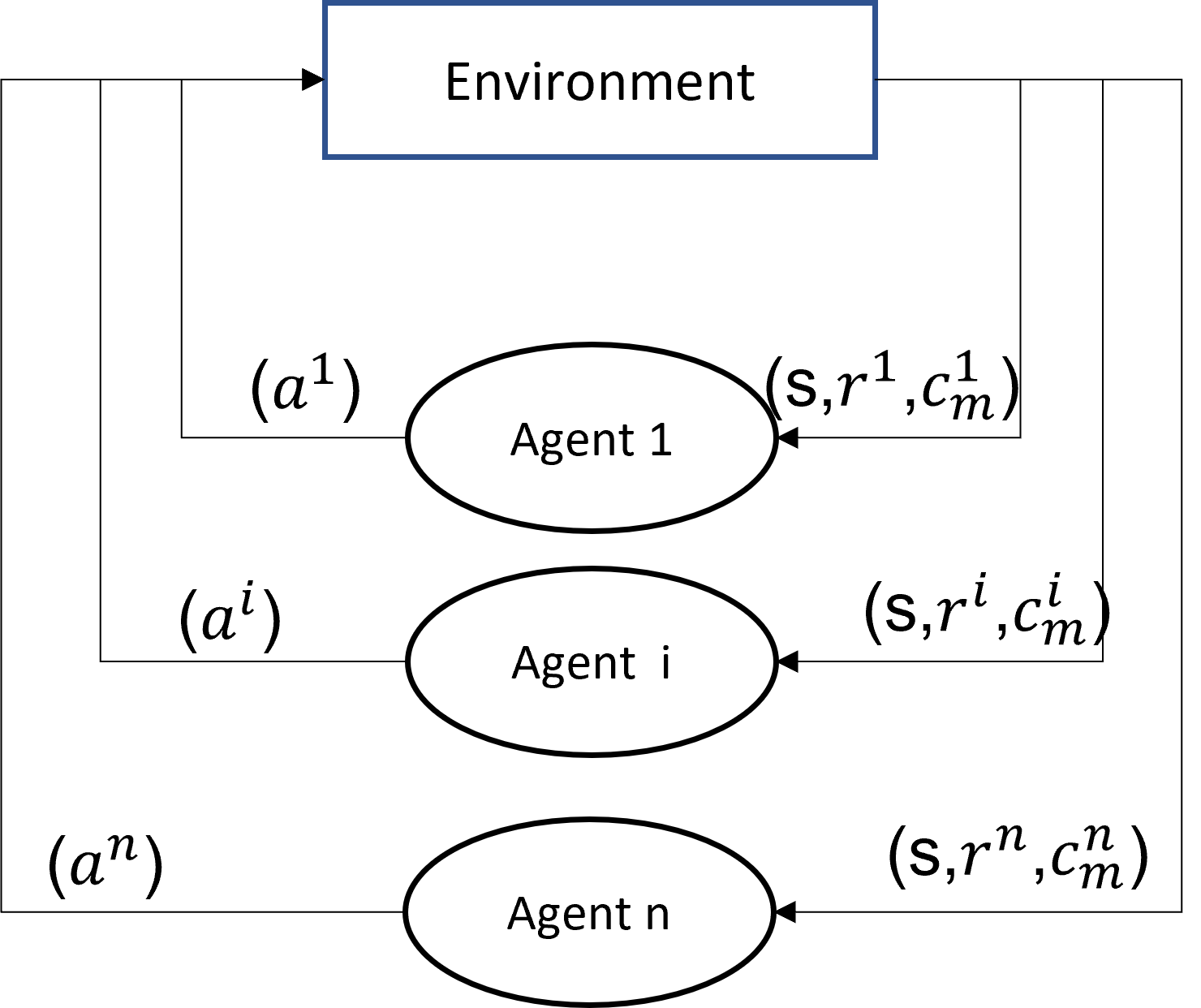}
    \caption{ Schematic diagrams of constrained Markov game, which correspond to multi-agent RL with safety constraints. Note that for every agent, there can be $m$ constraints, for simplicity, we assume there is only one constraint when $m=1$. } 
    \label{fig: cmarl}
\end{figure}

Conventional MARL does not explicitly account for safety, which is critical in domains where safety is of paramount importance. To address this, safe reinforcement learning \citep{achiam2017constrained,gu2024review, zheng2024safe} extends MARL by integrating safety constraints into the decision-making process (Figure \ref{fig: cmarl}). Each agent receives not only \textbf{rewards $r^{i}$} but also \textbf{penalties or costs $c^{i}$} when actions or resulting system states violate predefined safety thresholds. This constrained Markov game \citep{altman2021constrained} formulation restricts agents to a feasible action space that ensures compliance with operational and regulatory safety requirements -- in the case of CO$_{2}$ storage, limiting reservoir pressure to prevent caprock fracturing. By embedding safety directly into the learning process, this framework ensures that optimal strategies are not only economically beneficial but also operationally viable and environmentally secure.

A safe multi-agent reinforcement learning framework tailored for GCS is displayed in figure \ref{fig: marl_framework}. Each operator/agent is interacting with the subsurface environment and receiving reward $r^{i}$ and cost $c^{i}$ at each timestep. Based on the observation $o^{i}$, each operator chooses their own well control strategy $a^{i}$ and the joint actions of all operators {$a^{1}$, $a^{i}$, ..., $a_{n}$} drive the current environment state $s_{t}$ to next environment state $s_{t+1}$. In field operations, injection controls are typically updated at discrete management intervals rather than continuously. Accordingly, in our formulation each action $a_t$ represents a piecewise-constant well-control decision applied over a user-defined control period $\Delta t$. This abstraction is intended to reflect practical operational scheduling while retaining a sequential decision-making structure suitable for MDP/Markov-game formulations.

To incorporate \textbf{safety considerations}, the Markov game is extended into a 
\textbf{Constrained Markov Game (CMG)}:
\[
\langle n, \mathcal{S}, \{\mathcal{A}^i\}_{i=1}^n, \mathbb{T}, \{r^i\}_{i=1}^n, \{c^i\}_{i=1}^n, \gamma \rangle,
\]

where each agent additionally has a cost function $c^i(s, a)$ representing safety violations (e.g., exceeding reservoir pressure limits). Building upon this framework, the problem can be formulated as in equation \ref{formula: cmarl_obj}. In similar vein, 

\begin{equation}
Q_c^{i}(s,a^{i},a^{-i}) := \mathbb{E}\!\left[
\sum_{t=0}^{T}\gamma^{t} c_{t}^{i}
\;\middle|\;
s_{0}=s,\; a_{0}^{i}=a^{i},\; a_{0}^{-i}=a^{-i},\; a_{t}^{i}\sim \pi^{i}(\cdot\mid s_{t}),\; a_{t}^{-i}\sim \pi^{-i}(\cdot\mid s_{t})
\right].
\end{equation}

estimates the expected constraints violations. 






A common approach to solve this problem uses \textbf{Lagrangian relaxation}, 
where the objective is reformulated as:

\begin{equation}
\mathcal{L}^i(\pi, \lambda^i) 
= J^i(\pi) + \lambda^i \left( \mathbb{E}\left[ \sum_{t=0}^{\infty} \gamma_t c^i_t \right] - d^i \right),
\end{equation}

with $\lambda^i \geq 0$ being the Lagrange multiplier balancing reward maximization 
and constraint enforcement. This coefficient is dynamically updated to penalize constraint violations.

\section{Proposed Framework - Constrained Markov Game for GCS}
\label{sec:Methodo}

\subsection{Safe Multi-Agent DDPG (MADDPG) for GCS}

To ensure each operator injects safely while maximizing potentially competing objectives in basin-scale CCS, we adopt the Multi-Agent Deep Deterministic Policy Gradient (MADDPG) framework \citep{lowe2017multi}. This framework employs centralized training with decentralized execution (CTDE) to enable multiple operators—modeled as agents—to learn coordinated, well-control policies.

In this setup, the centralized critic network $Q_{r}^i(s,a^i,a^{-i})$ is used only during training to evaluate action values under the joint state--action information and to provide a stable learning signal; it does not
determine or execute actions. Each agent's actor network $\mu_{\theta^i}(o^i)$ outputs its own action based on local observations (e.g., pressures, plume saturation) . The economic objective optimized by each agent depends on the coalition structure through the reward formulation: in the fully cooperative case, agents share a common team reward (total NPV), whereas in the competitive (or mixed) cases each agent maximizes its own expected discounted economic return. The complete set of coalition structures is summarized in Table \ref{table:coalition structures}. 

In this study, each operator's observation is restricted to an operator-centric set (e.g., bottom-hole pressure and locally available variables) to reflect limited data availability in GCS settings. Pressure interference from neighboring projects is still captured through the shared reservoir dynamics and is reflected in each operator's measurements and constraint costs. If boundary or regional pressure monitoring is available or mandated (e.g., observation wells near permit boundaries or regulator-provided measurements), these data can be incorporated directly into the observation vector $o_i$ as additional channels without changing the proposed formulation.

 



\begin{figure}[htbp]
    \centering
    \includegraphics[width=0.6\textwidth]{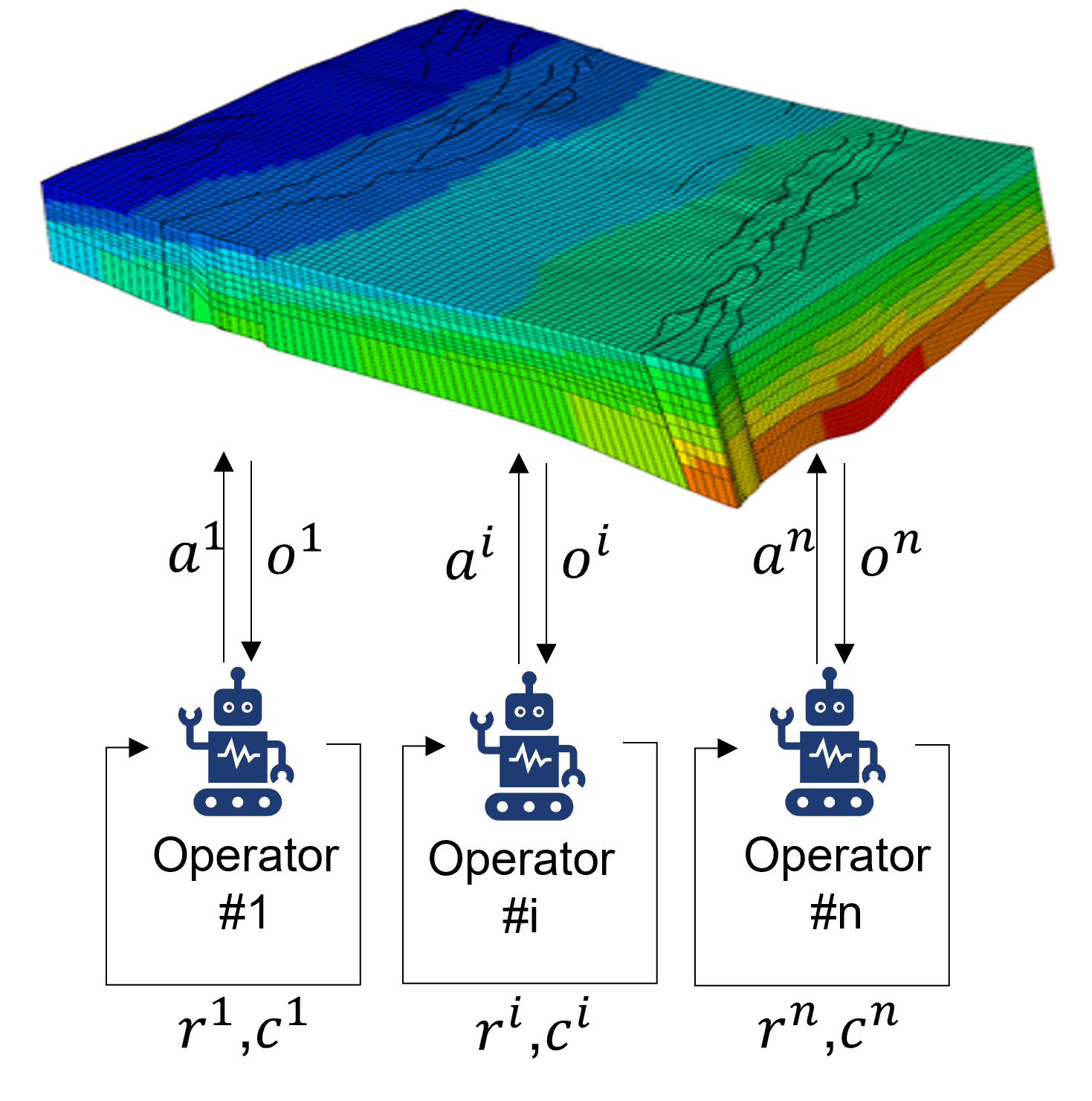}
    \caption{ A safe multi-agent reinforcement learning framework tailored for multi-stakeholder CCS projects. Each agent/operator is interacting with subsurface dynamical environment, receiving immediate reward (defined by present values) and penalty (e.g. injection pressure exceeding the fracturing pressure) and driving the environment to next state.} 
    \label{fig: marl_framework}
\end{figure}

\subsection{Adaptations for GCS Optimization}

The standard MADDPG framework is adapted to account for the unique features of CCS operations:

\begin{itemize}
    \item \textbf{Reduced-order surrogate modeling:} 
    A machine learning-based proxy, built on the Embed-to-Control (E2C) framework, 
    is employed to approximate high-dimensional reservoir dynamics. 
    This significantly reduces computational overhead while preserving accuracy.

    \item \textbf{Compliance-aware learning:} 
    Safety is enforced by incorporating a cost critic network $Q^i_{c}$ and a Lagrangian multiplier $\lambda^i$, ensuring that pore pressure at well locations or each grid cell remains below the allowable threshold pressure ($p_{\text{grid}} \leq \alpha p_{\text{frac}}$). Specifically, the actor network for agent $i$ is trained using the Lagrangian objective: 
    \begin{equation}
    \mathbb{E}\Big[-\,Q_r^{i}(s,a)\;+\;\lambda^i\,Q_c^{i}(s,a)\,\Big],
    \end{equation}
    where $\lambda^i \geq 0$ is a non-negative weight that controls how strongly safety violations are penalized. It is automatically adjusted during training—if an agent is breaking the safety limit too often, $\lambda^i$ increases to apply more penalty. The critics are learned through temporal-difference (TD) regression with experience replay and target networks; the actors are updated via deterministic policy gradients; and target networks are soft-updated to stabilize learning. The detailed training and execution scheme was displayed in Figure \ref{fig: maddpg_scheme} and the algorithm is shown in Algorithm \ref{alg:cmaddpg}. 
\end{itemize}

It should be noted that the pressure limit used in this work is an illustrative and widely adopted safety constraint, but it may not be the binding constraint in every basin-scale setting (e.g., highly open boundaries or low injection intensity). The proposed constrained Markov game/safe-MARL framework is not restricted to pressure: other operational or regulatory constraints (e.g., plume extent/area of review (AoR), leakage-risk indicators) can be incorporated by defining the agent-specific cost function $c_i(s,a)$ accordingly. Nevertheless, even at basin scale, pressure can become limiting when multiple operators inject concurrently and interference accumulates over time. 

\subsection{Reward and Penalty Formulation for Geological \texorpdfstring{CO$_2$}{CO2} Storage}

In reinforcement learning, a reward model defines the feedback an agent receives for its actions, signaling which behaviors are beneficial and guiding the agent toward desired outcomes \citep{yu2025reward, arora2021survey}. In safe reinforcement learning, this is complemented by a penalty model, which assigns negative feedback when actions violate safety constraints. Together, reward and penalty models ensure that agents not only maximize designated performance but also operate within required safety and risk boundaries. In multi-agent reinforcement learning (MARL) for geological CO$_2$ storage (GCS), rewards and penalties must capture the trade-offs between economic returns and regulatory compliance, while accounting for the dynamic, time-dependent nature of subsurface processes.

Common MARL frameworks classify reward structures into three categories: fully competitive, fully cooperative, or mixed competitive–cooperative. However, GCS presents added complexity due to evolving subsurface conditions. At early injection stages, when pressure space is abundant, agents may independently maximize their own net present values (NPVs) without adversely affecting others. At later stages, as pore pressure space becomes constrained, one operator’s increased injection may reduce the feasible capacity of others, introducing significant interdependence and potential conflicts.

To address these challenges, we adopt coalition structures inspired by game theory, enabling flexible combinations of cooperative and competitive behaviors. Coalition structures are enumerated using Bell numbers, allowing evaluation of all possible collaborative configurations. In our case study with three operators (A, B, and C), 5 different coalition structures are considered: the coalition types, structures, and corresponding objectives and reward formulations are summarized in Table \ref{table:coalition structures}. The fully cooperative one is where all companies act collectively to maximize the total net present value (NPV) of the project, and fully competitive one is where each company independently seeks to maximize its own NPV without regard for the others. This framework allows us to capture a wide spectrum of strategic behaviors, from joint optimization for mutual benefit to self-interested competition. Based on these coalition types, distinct reward and penalty structures are defined to reflect the corresponding operational incentives and constraints. 

\begin{table}[h!]
\centering
\caption{Coalition types, structures, objectives, and corresponding reward formulations.}
\resizebox{\textwidth}{!}{
\begin{tabular}{p{4cm} p{4cm} p{6cm} p{6cm}}
\hline
\textbf{Coalition Types} & \textbf{Coalition Structure} & \textbf{Objective} & \textbf{Reward Definition}\\
\hline

Fully competitive
& \textbf{\{A\},\{B\},\{C\}} 
& Each agent $i$ acts independently and maximizes its own financial return. 
& $r^{A}=NPV^{A},\; r^{B}=NPV^{B},\; r^{C}=NPV^{C}$ \\

Fully cooperative/collaborative
& \textbf{\{A \& B \& C\}} 
& Agents A, B, and C form a grand coalition to maximize the collective financial outcome of the team.
& $r^{A}=r^{B}=r^{C}=NPV^{A}+NPV^{B}+NPV^{C}$  \\

Partially cooperative: coalition AB vs. C
& \textbf{\{A \& B\}, \{C\}}
& Agents A and B cooperate to maximize their joint return, while agent C acts independently to maximize its own NPV.
& $r^{A}=r^{B}=NPV^{A}+NPV^{B},\; r^{C}=NPV^{C}$  \\

Partially cooperative: coalition AC vs. B
& \textbf{\{A \& C\}, \{B\}}
& Agents A and C cooperate to maximize their coalition reward, while agent B maximizes its own individual return.
& $r^{A}=r^{C}=NPV^{A}+NPV^{C},\; r^{B}=NPV^{B}$   \\

Partially cooperative: coalition BC vs. A
& \textbf{\{B \& C\}, \{A\}}
& Agents B and C act as a coalition to improve their combined NPV, whereas agent A optimizes individually.
& $r^{B}=r^{C}=NPV^{B}+NPV^{C},\; r^{A}=NPV^{A}$ \\

\hline
\end{tabular}
}
\label{table:coalition structures}
\end{table}



For the penalty formulation, the incurred cost is defined to be proportional to the amount by which the pore pressure in the injection-well grid block exceeds a prescribed threshold:

\begin{equation}
\label{formula: penalty_well}
c^i_t = 5000 \cdot 
\sum_{g \in \mathcal{G}_{well}^i} 
\mathbf{1}\!\left( p_g^t > p_{thr} \right),
\end{equation}


In this formulation, $c^i_t$ denotes the penalty/cost incurred by agent $i$ at time step $t$. 
The set $\mathcal{G}^i$ represents all grid blocks within company $i$'s project area, and 
$p_g^t$ is the pressure in grid block $g$ at time $t$. The indicator function 
$\mathbf{1}(p_g^t >= p^i_{\text{thr}})$ returns $1$ if the pressure in injection-well grid block $g$ 
exceeds the threshold pressure $p^i_{\text{thr}}$ assigned to agent $i$, and $0$ otherwise. 
Thus, the summation counts the total number of violating injection-well grid blocks, which is then 
multiplied by a fixed penalty factor of $5000$ (5000 is selected in a way that the penalty value is of same magnitude of the reward in order for good training practice) to determine the agent's cost at that time step.

This penalty is imposed to train the constrained MADDPG to restrict the actions to feasible solutions: 

\[
\sum_{t=1}^{T} \gamma_t \cdot c^i(t) \leq d^i
\]

$d^i$ is the penalty threshold, in this experiment, because the pressure constraints are strictly imposed, $d^i$ is set to be 0.0.

This coalition-based reward structuring reflects realistic CCS scenarios, where some stakeholders may choose to collaborate for shared economic and safety benefits, while others pursue individual interests. By embedding coalition flexibility within the MARL framework, we capture the dynamic interplay between cooperation and competition inherent to multi-stakeholder CCS projects.

\subsection{Overall Workflow}
The overall workflow begins with initializing both the surrogate reservoir model and the multi-agent learning networks. During the training phase, each agent interacts with the reduced-order reservoir environment, generating state–action trajectories that include both reward and penalty signals. These interactions reflect the operational decisions of individual stakeholders and their resulting economic outcomes and constraint violations.

All generated trajectories are stored in a replay buffer, from which minibatches are sampled to update the actor and critic networks through gradient-based optimization. In parallel, Lagrangian multipliers are updated iteratively to enforce the defined safety constraints, adapting dynamically to the observed level of constraint violations. This coupled update process ensures that the learned policies not only maximize economic returns but also remain compliant with operational safety requirements.

Upon convergence, the trained policies are deployed in a decentralized manner, where each operator controls its own wells exclusively based on local observations without access to global state information. This structure mirrors realistic CCS operational settings, where information sharing between companies may be limited or restricted. The complete implementation of this learning and control process is formalized in Algorithm~\ref{alg:cmaddpg}, which integrates the reward and penalty critics, policy updates, and Lagrangian dual optimization into a unified safe multi-agent decision-making framework.

\section{Case Study and Results}
\label{sec:Results}

\subsection{Description of Reservoir and Model Setup}
\label{subsec:model_setup}
\begin{figure}[htbp]
    \centering
    \includegraphics[width =\textwidth]{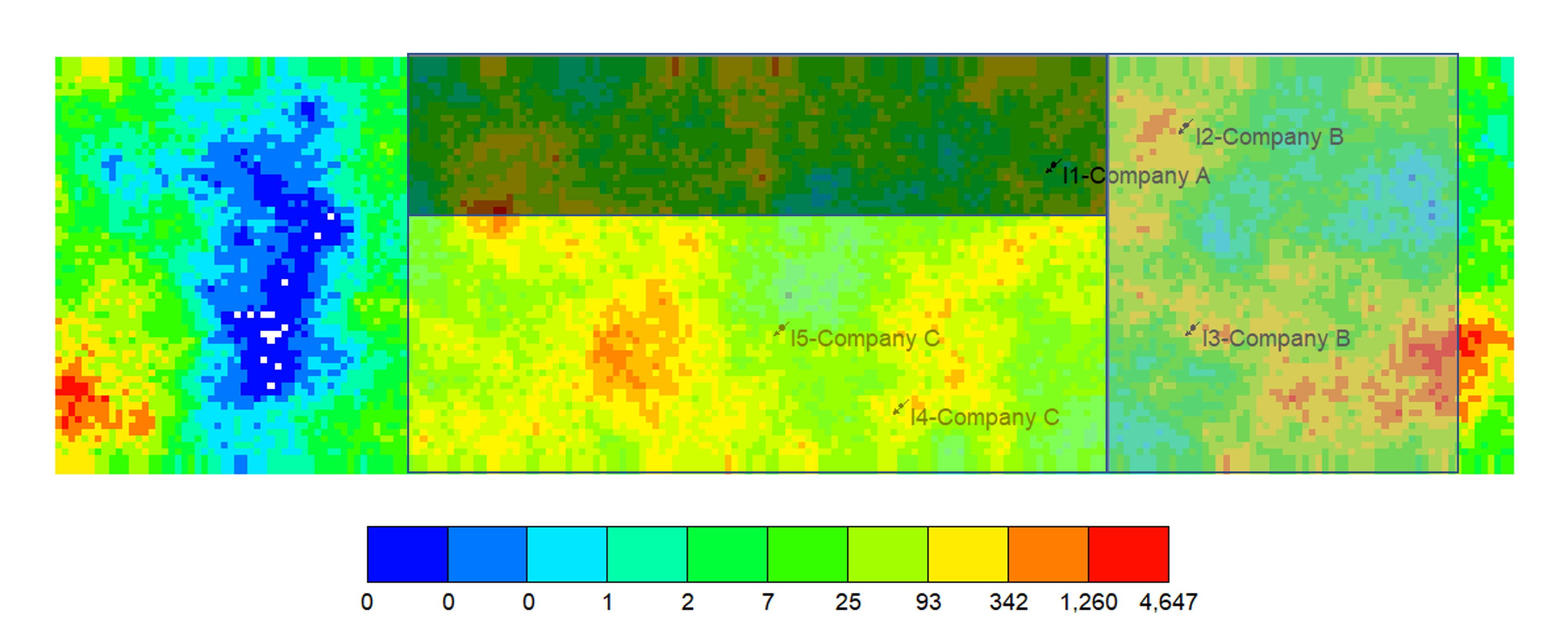}
    \caption{Permeability map (in milliDarcy) of the synthetic basin area. Company A, B and C are operating different numbers of injection/extraction wells. The shadowed area indicating the leased project area of each company.} 
    \label{fig: perm}
\end{figure}

The proposed framework is implemented in a reservoir where the dimension spans $44 \mathrm{km} \times 12.8 \mathrm{km} \times 0.2\mathrm{km}$ in x, y and z directions. 
With a depth of $4{,}500 \ \mathrm{m}$ and an assumed fracture gradient of $17.0 \ \mathrm{kPa/m}$, the fracture pressure $p_{\mathrm{frac}}$ is set to be $76{,}500 \ \mathrm{kPa}$. For safe CO$_2$ injection, the pore pressure must remain below the fracturing pressure, expressed as:
\[
p_{\mathrm{grid}} \leq \alpha \, p_{\mathrm{frac}},
\]
where $0.0 < \alpha < 1.0$ reflects site-specific risk tolerance. In this study, $\alpha$ should also account for the prediction error between the reduced-order proxy model and the high-fidelity model. The permeability distribution for the basin is shown in Figure~\ref{fig: perm}.

We consider three companies (A, B, and C) operating in non-overlapping regions of the basin (as indicated in the shaded area in Figure \ref{fig: perm}), each managing a different number of injection wells. To reflect varying risk tolerances, the maximum allowable pore pressure for each company is set to:
\[
p_{\mathrm{threshold},i} = \alpha_i \, p_{\mathrm{frac}} =
\begin{cases}
75{,}000 \ \mathrm{kPa}, & i = A, \\
65{,}000 \ \mathrm{kPa}, & i = B, \\
65{,}000 \ \mathrm{kPa}, & i = C.
\end{cases}
\]
The initial reservoir pressure $p_{\mathrm{init}}$ is $20{,}000 \ \mathrm{kPa}$. All injection wells are constrained to operate between a minimum rate of $0.5 \ \mathrm{MMTon/year}$ and a maximum rate of $5.0 \ \mathrm{MMTon/year}$:
\[
0.5 \ \mathrm{MMTon/year} \ \le q_{\mathrm{inj}} \le \ 5.0 \ \mathrm{MMTon/year}.
\]
Economic assumptions for revenue and cost calculations are summarized in Table~\ref{tab:revenue&cost}, with all companies sharing the same unit revenue and cost structure. Injection rate of each company is varied every 1 year and the total injection lasts 20 years. 



\begin{table}[htbp]
\centering
\caption{CCS project revenues and costs}
\begin{tabular}{|l|c|}
\hline
\textbf{Revenues/costs} & \textbf{Value} \\
\hline
Tax credit ($R_{\text{credit}}$), \$/ton & 85 \\
\hline
CO\textsubscript{2} capture, transportation and storage cost ($R_{\text{op}}$), \$/ton & 45 \\
\hline
Water disposal cost ($R_{\text{w}}$), \$/ton & 30 \\
\hline
Discount rate $\gamma$ & 0.95 \\
\hline
\end{tabular}
\label{tab:revenue&cost}
\end{table}


\subsection{Fully Cooperative vs. Fully Competitive}
\begin{figure}[htbp]
    \centering
    \begin{minipage}[t]{0.48\textwidth}
        \centering
        \includegraphics[width=\textwidth]{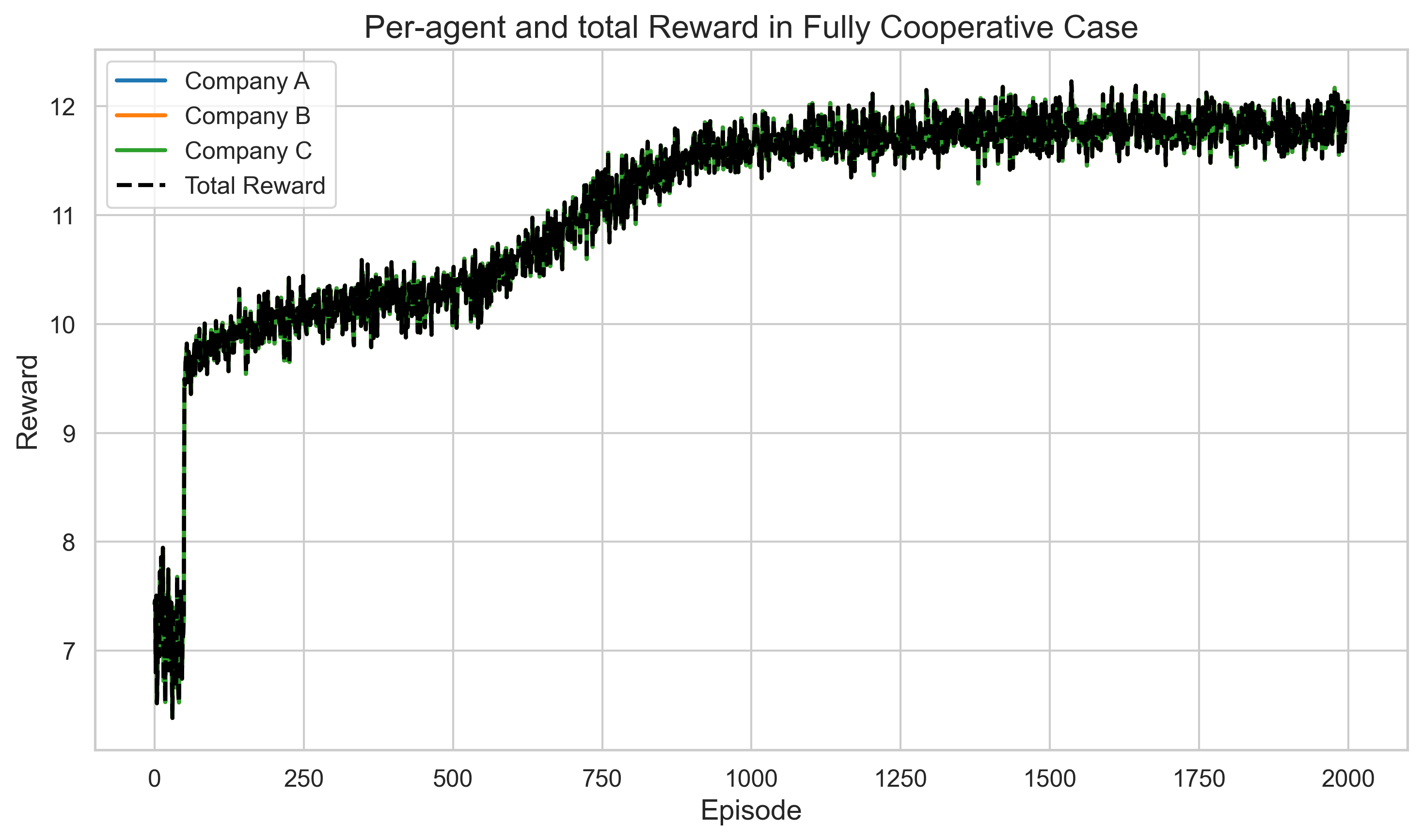}
        \vspace{0.5em}
        \small (a) Per-agent and total reward in fully collaborative scenario
    \end{minipage}
    \hfill
    \begin{minipage}[t]{0.48\textwidth}
        \centering
        \includegraphics[width=\textwidth]{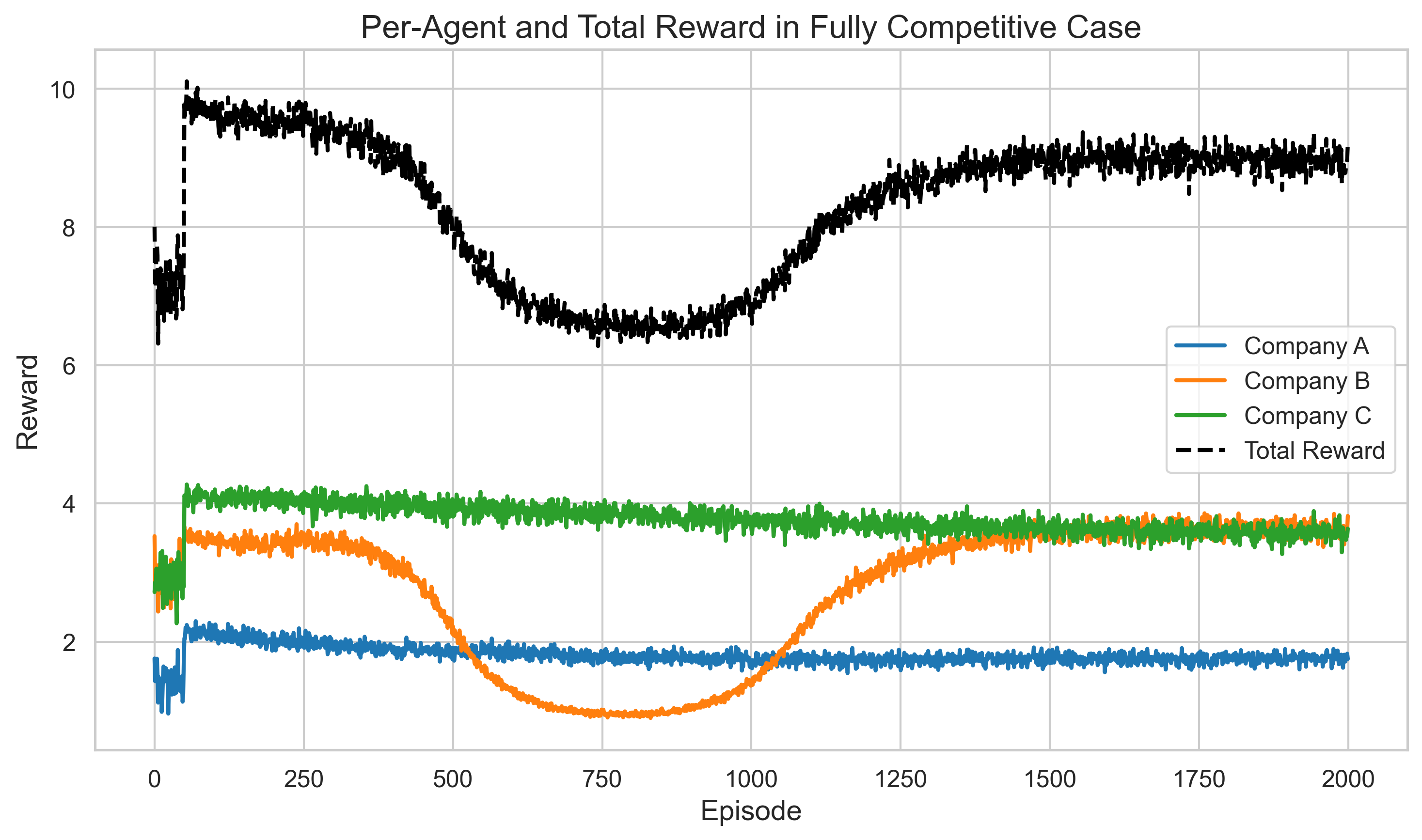}
        \vspace{0.5em}
        \small (b) Per-agent and total reward in fully competitive scenario
    \end{minipage}
    \vspace{1em} 
    \begin{minipage}[t]{0.48\textwidth}
        \centering
        \includegraphics[width=\textwidth]{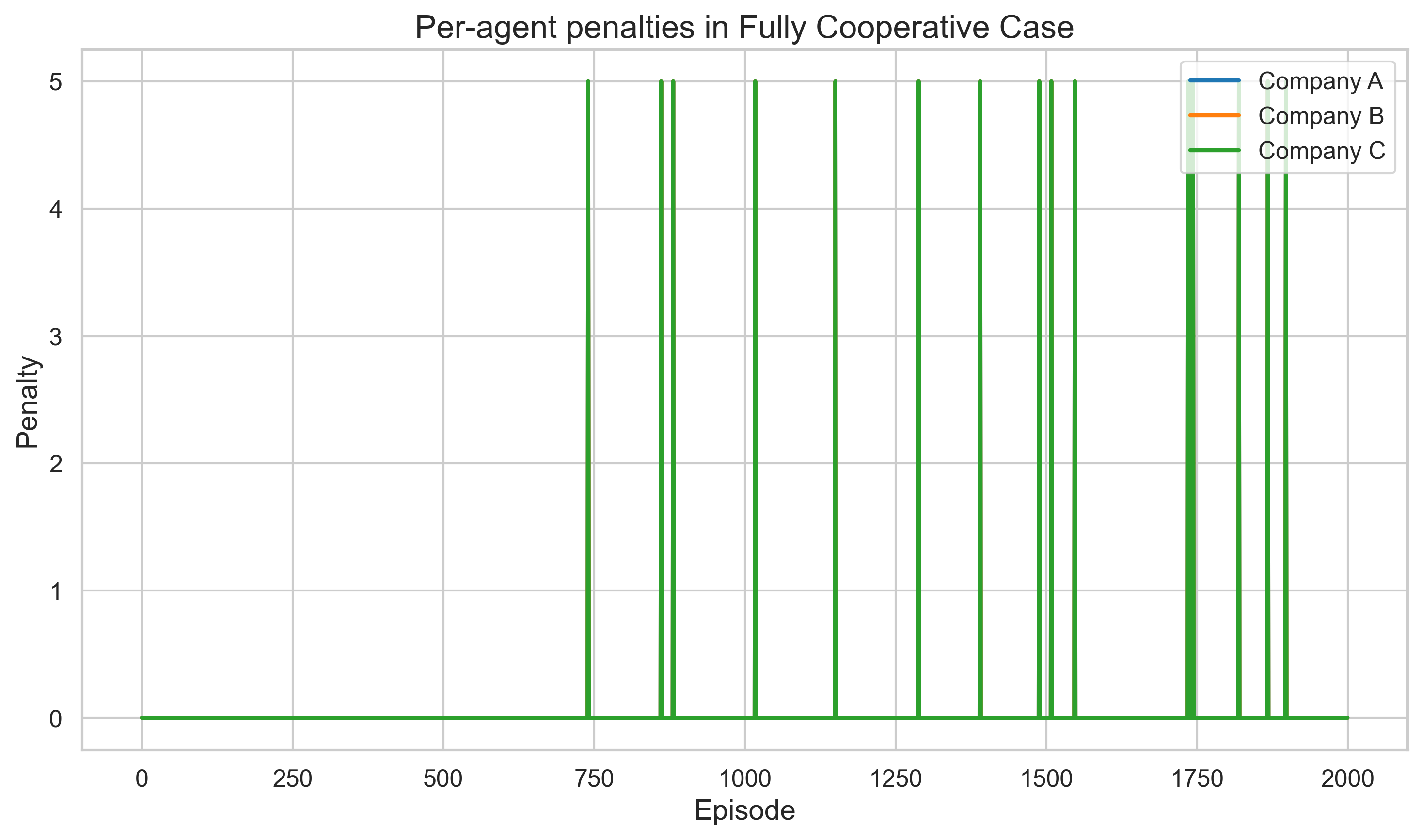}
        \vspace{0.5em}
        \small (c) Per-agent penalty in fully collaborative scenario
    \end{minipage}
    \hfill
    \begin{minipage}[t]{0.48\textwidth}
        \centering
        \includegraphics[width=\textwidth]{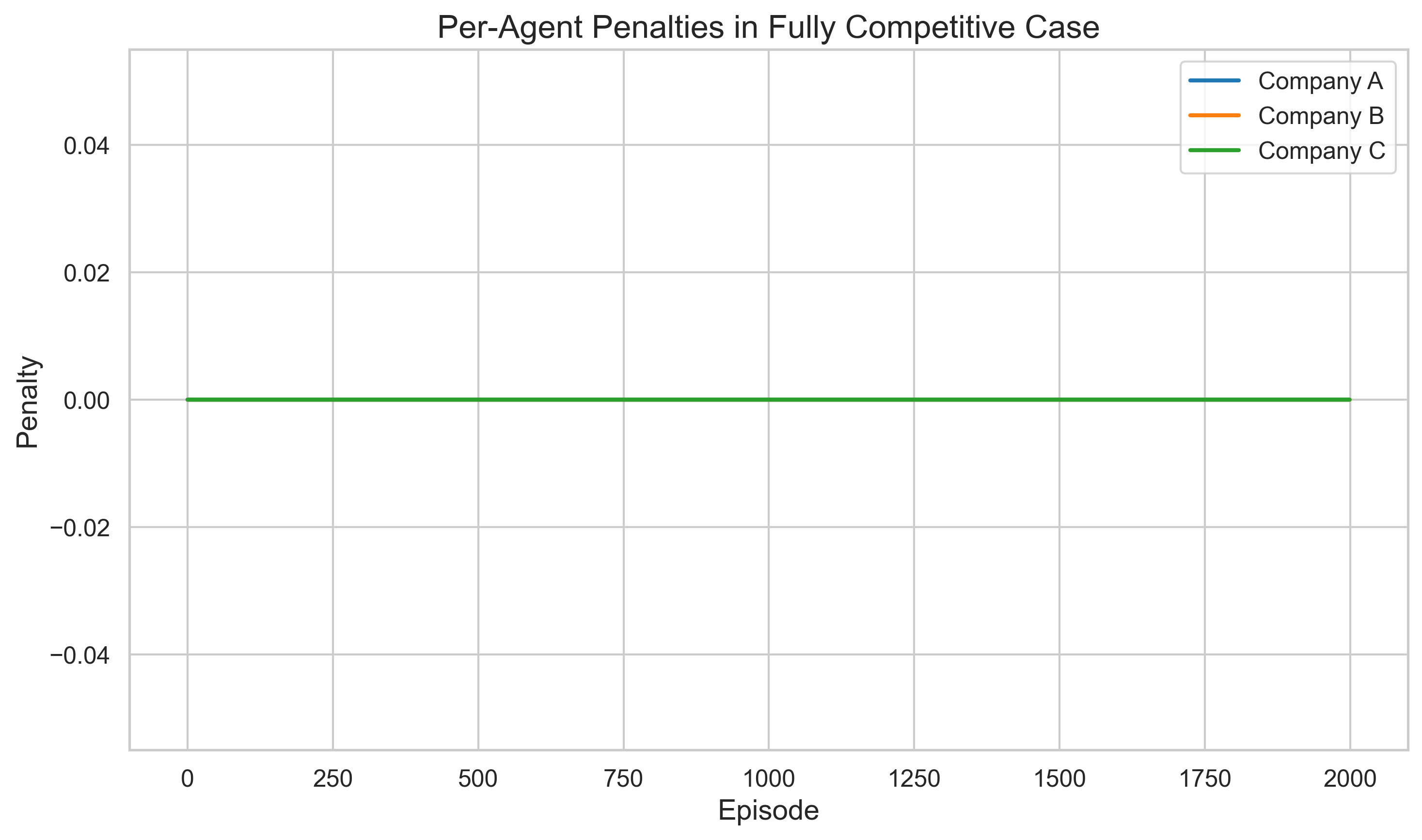}
        \vspace{0.5em}
        \small (d) Per-agent penalty in fully competitive scenario
    \end{minipage}

    \caption{Per-agent and total reward during training for (a) the fully cooperative scenario and (b) the fully competitive scenario, along with the corresponding per-agent penalties for (c) the fully cooperative scenario and (d) the fully competitive scenario. Each episode represents a complete simulation run of the CO$_2$ storage over the defined time horizon. The episode index reflects training progress: early episodes exhibit more exploratory behavior, whereas later episodes correspond to increasingly optimized policies.}
    \label{fig:merged_reward-train}
\end{figure}

In this subsection, we present two baseline scenarios: a fully cooperative case in which all companies form a grand coalition to maximize total team NPV, and a fully competitive case where each company operate independently without any coalition (fully competitive case). Figure \ref{fig:merged_reward-train} displays the reward and penalty trajectories over the training process.

In the fully cooperative case, all companies share the same team reward, see Table \ref{table:coalition structures}. As observed from Figure \ref{fig:merged_reward-train} (a), it starts with conservative injection operations and improves steadily up to approximately episode 1000, after which the reward curve levels off, indicating convergence towards optimal policies. During early training, the policy may temporarily violate the safety threshold (as reflected by spikes in the penalty curve \ref{fig:merged_reward-train} (c)), but it ultimately stabilizes at a solution that satisfies the safety requirement, demonstrating constraint-compliant operation.

In the fully competitive case, each company also starts conservatively and rapidly improves its own policy during the first 50 episodes. Between roughly episodes 50 and 750, companies adapt to one another’s actions and competitive responses intensify, which reduces the total reward. From about episode 750 to 1500, the policies adjust toward a more balanced outcome and the total reward recovers, eventually stabilizing after episode 1500, consistent with convergence to an equilibrium-like behavior.

\begin{figure}[htbp]
    \centering
    \begin{minipage}[t]{0.48\textwidth}
        \centering
        \includegraphics[width=\textwidth]{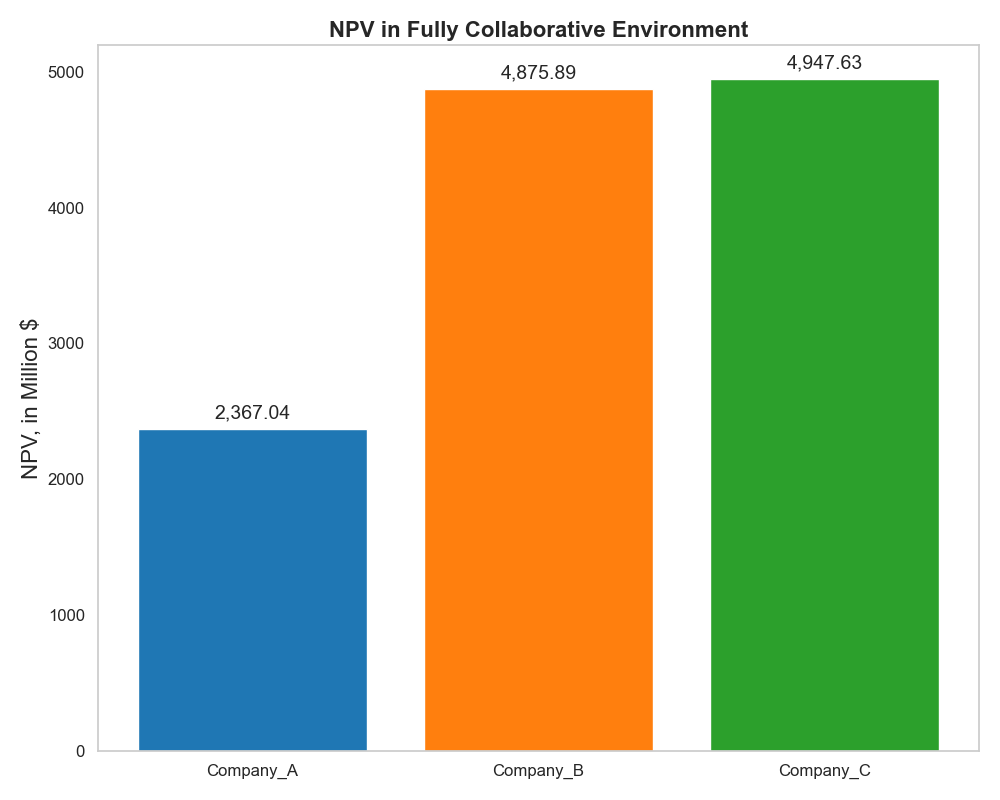}
        \vspace{0.5em}
        \small (a) NPVs in cooperative setting
    \end{minipage}
    \hfill
    \begin{minipage}[t]{0.48\textwidth}
        \centering
        \includegraphics[width=\textwidth]{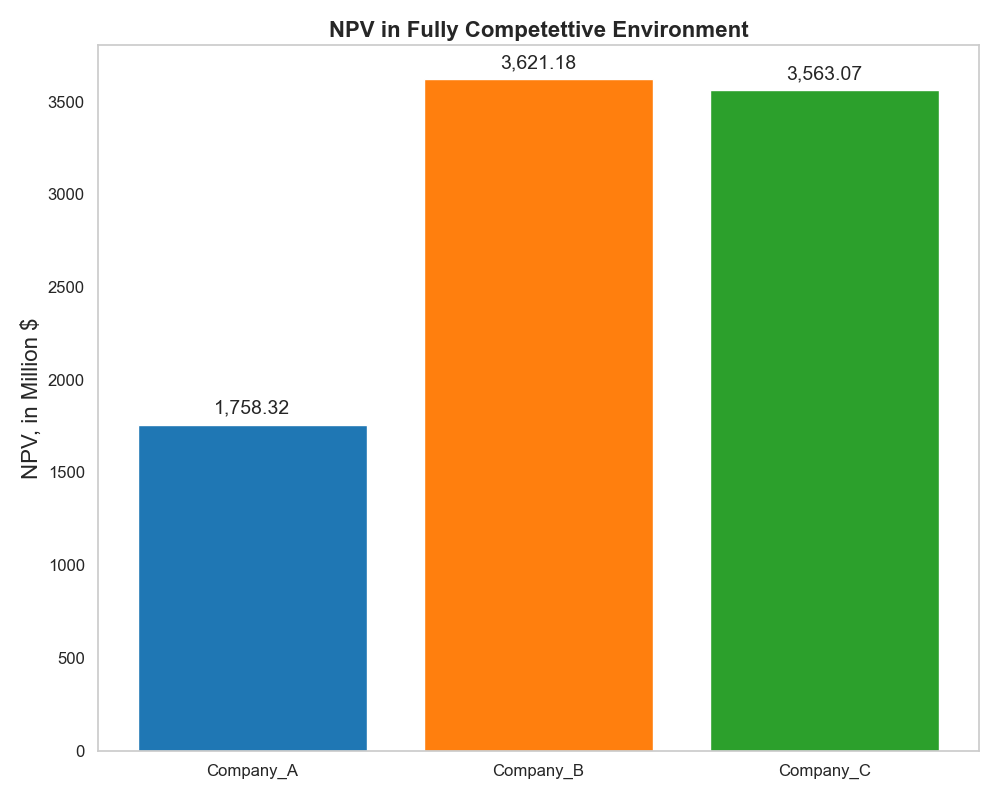}
        \vspace{0.5em}
        \small (b) NPVs in competitive setting
    \end{minipage}

    \caption{NPV of individual company under (a) fully collaborative scenario and (b) fully competitive scenario. The total team NPV is is \$12191 Million for the fully cooperative case and \$8942 Million for the fully competitive case..}
    \label{fig:merged_reward_hist}
\end{figure}

Figure \ref{fig:merged_reward_hist} summarizes the resulting NPVs under these two settings. The total team NPV increases from \$8942 million in the competitive setting to \$12191 million in the cooperative setting. At the company level, all three companies gain substantial benefits from coalition formation, with NPV gains of \$609 million, \$1255 million and \$1385 million respectively, compared to acting alone. To that end, there are clear incentives for all operators to participate in the grand coalition and to coordinate the operations in the basin.

 
\begin{figure}[htbp]
    \centering
    \begin{minipage}[t]{0.48\textwidth}
        \centering
        \includegraphics[width=1.2\textwidth]{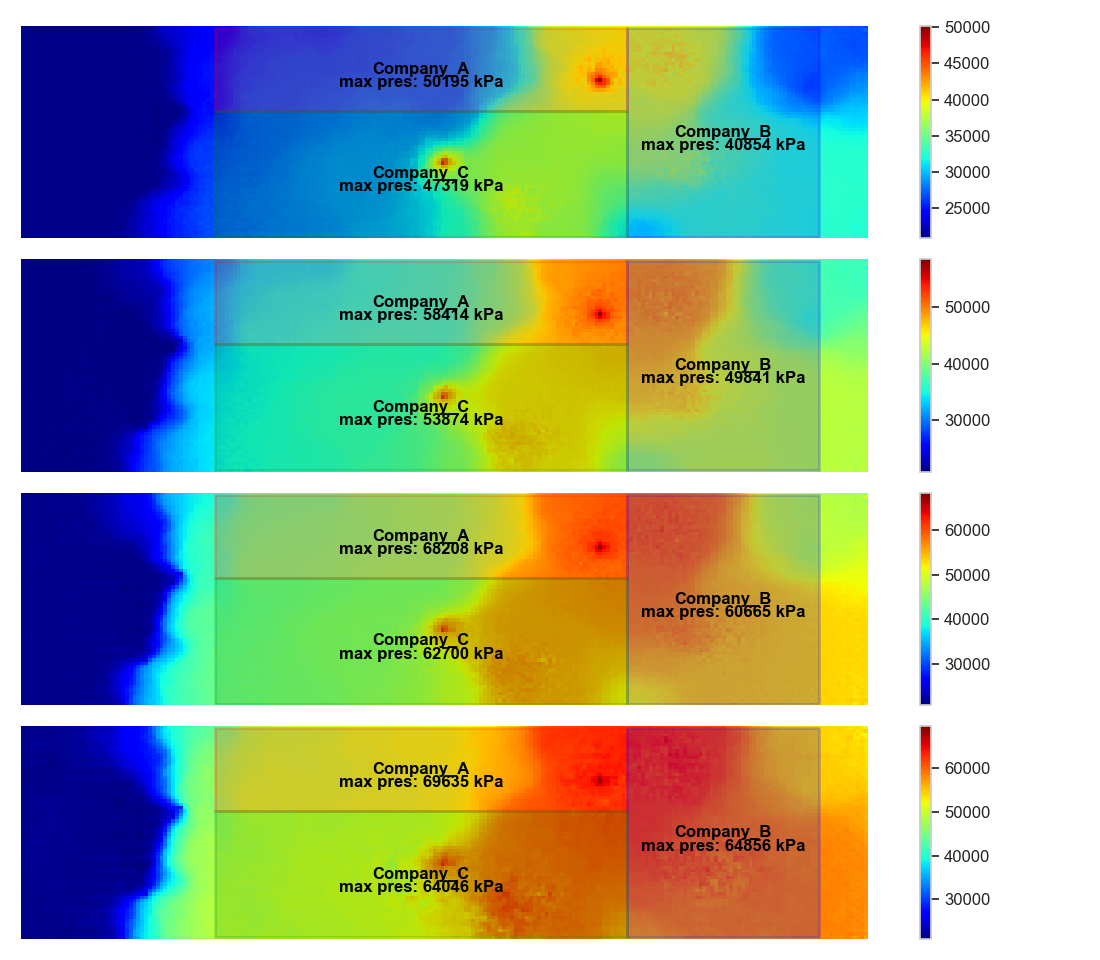}
        \vspace{0.5em}
        \small (a) Pressure map (fully cooperative) at selected injection periods, in kPa
    \end{minipage}
    \hfill
    \begin{minipage}[t]{0.48\textwidth}
        \centering
        \includegraphics[width=1.2\textwidth]{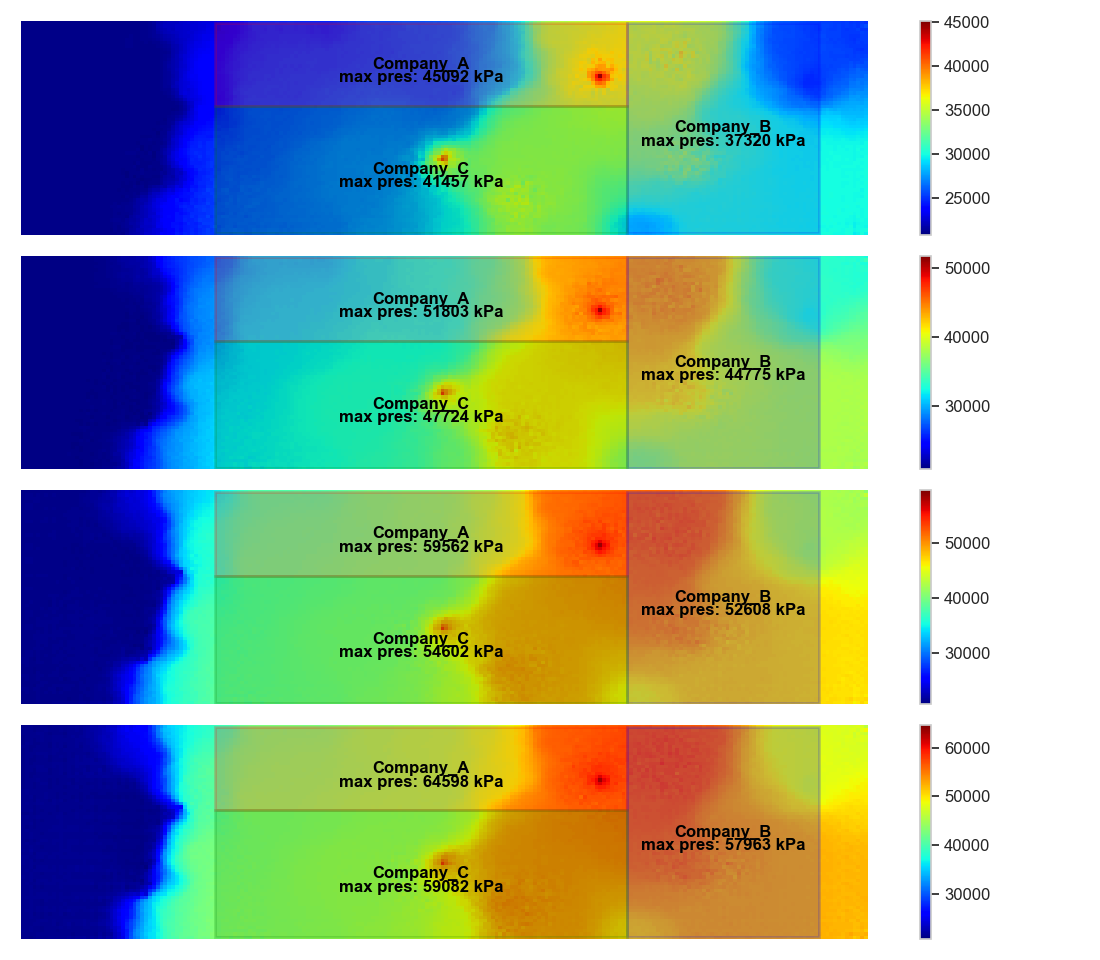}
        \vspace{0.5em}
        \small (b) Pressure map (fully competitive) at selected injection periods, in kPa
    \end{minipage}

    \caption{Pressure maps at different injection stages t = 5, 10, 15, 20 years (from top to bottom) for (a) fully collaborative scenario and (b) fully competitive scenario. The maximum pressure reached at each leased project area is annotated.}
    \label{fig:merged_pressure_maps}
\end{figure}

Overall, the proposed framework maintains injection pressures within safe limits throughout the injection period, ensuring compliance with operational safety requirements.. As shown in Figure \ref{fig:merged_pressure_maps}, the maximum pressures at injection-well locations remain below the fracturing pressure at the end of injection in both the fully cooperative and fully competitive settings. Notably, grand coordination enables substantially better utilization of the available pressure space, particularly for Company B, highlighting the effectiveness of the proposed approach. In contrast, under full competition, self-interested responses to opponents’ actions prevent efficient use of the available pressure space.

\subsection{Mixed Cooperative and Competitive}

\begin{figure}[htbp]
    \centering
    \begin{minipage}[t]{0.48\textwidth}
        \centering
        \includegraphics[width=\textwidth]{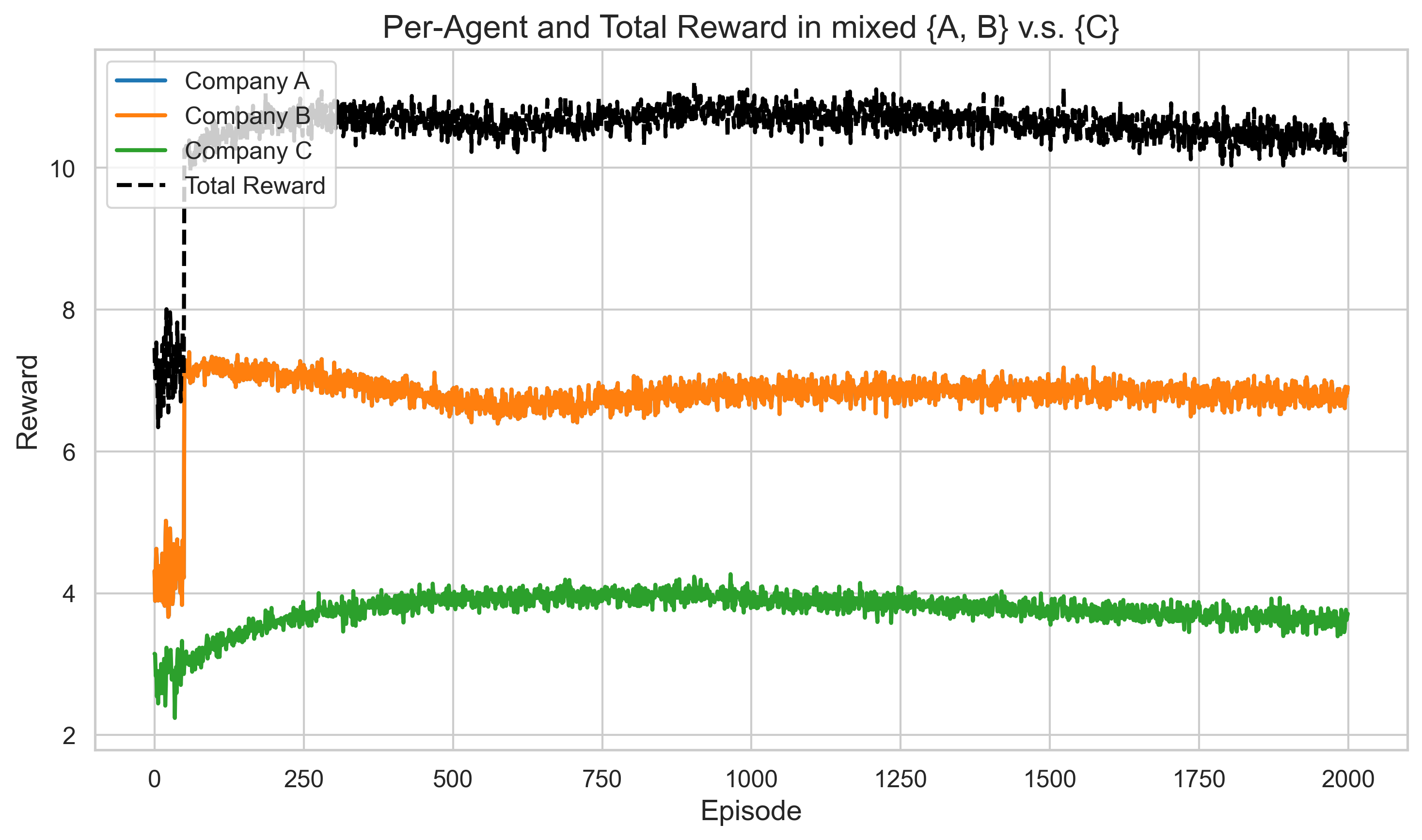}
        \vspace{0.5em}
        \small (a) Per-agent and total reward in mixed \{A, B\} and \{C\} scenario
    \end{minipage}
    \hfill
    \begin{minipage}[t]{0.48\textwidth}
        \centering
        \includegraphics[width=\textwidth]{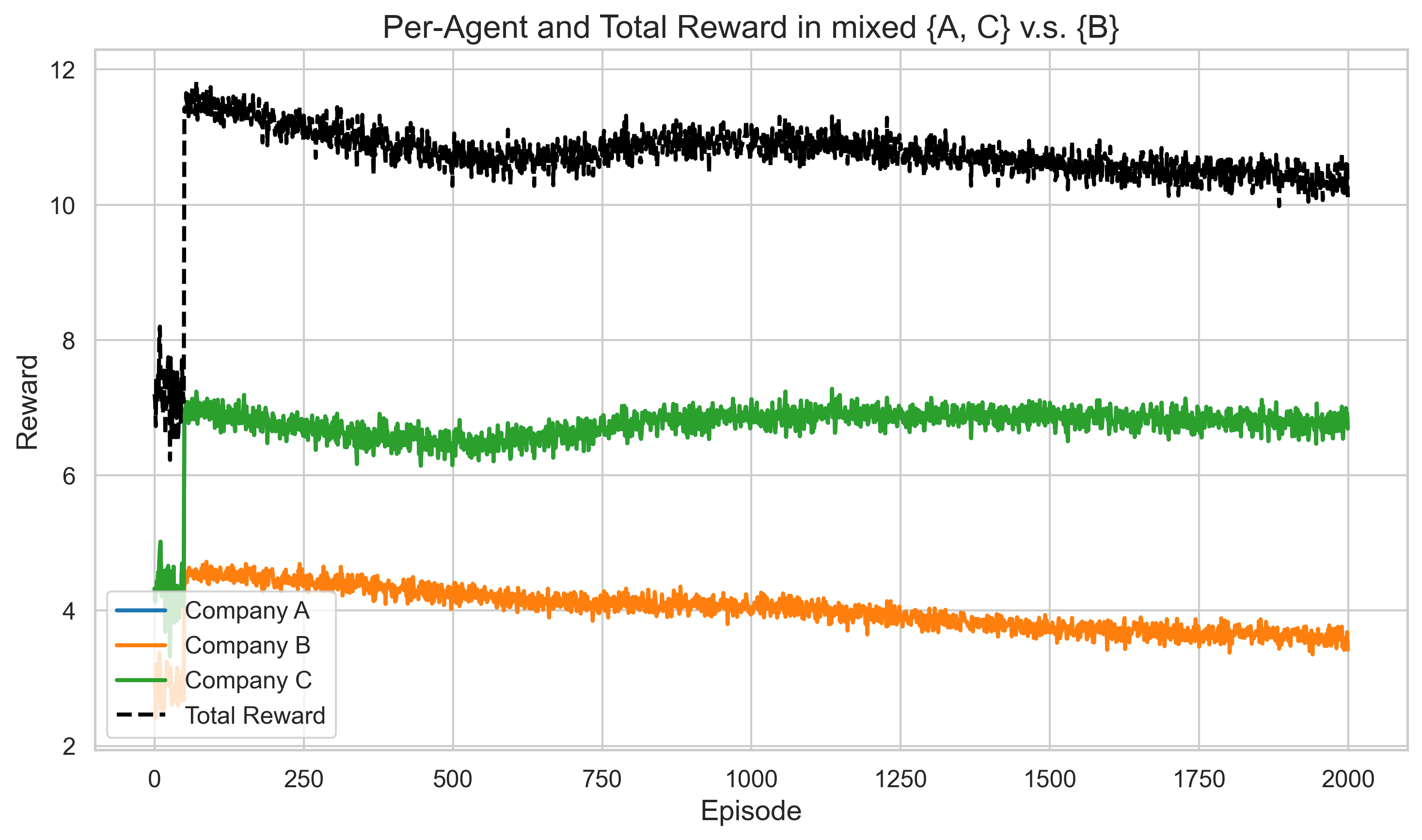}
        \vspace{0.5em}
        \small (b) Per-agent and total reward in mixed \{A, C\} and \{B\} scenario
    \end{minipage}
    \vspace{1em} 
    \begin{minipage}[t]{0.48\textwidth}
        \centering
        \includegraphics[width=\textwidth]{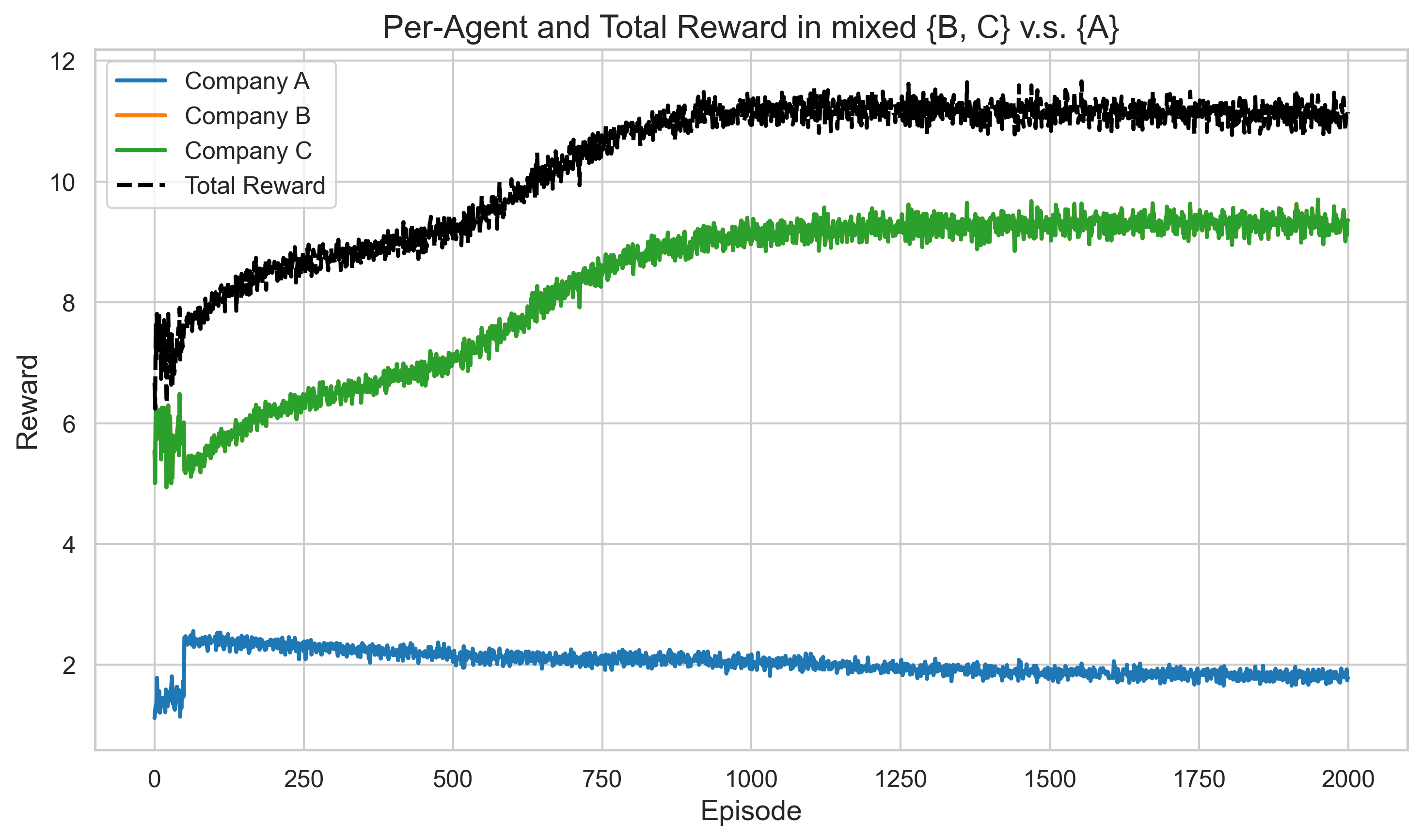}
        \vspace{0.5em}
        \small (c) Per-agent and total reward in mixed \{B, C\} and \{A\} scenario
    \end{minipage}
    \hfill

    \caption{Per-agent and total reward trajectories during training for the three mixed-coalition settings: (a) \{A, B\} v.s \{C\}, (b) \{A, C\} v.s \{B\} and (c) \{B, C\} v.s \{A\}. }
    \label{fig:merged_mixed_reward-train}
\end{figure}

In this subsection, we examine partially collaborative settings in which some companies cooperate while competing with the remaining ones. This leads to three coalition structures: (i) A and B cooperate while competing with C, (ii) A and C cooperate while competing with B, and (iii) B and C cooperate while competing with A. Figure \ref{fig:merged_mixed_reward-train} summarizes the corresponding reward trajectories during training.

Across all three mixed-coalition scenarios, a consistent pattern emerges that echoes the previous subsection: forming a coalition provides a systematic advantage over operating alone. When Companies A and B form a coalition against C, the coalition members quickly converge to a stable, relatively high reward, whereas the independent agent C settles into a lower but steady payoff as it adapts to the coalition’s coordinated behavior. A similar dynamic appears when A cooperates with C against B: the coalition partners achieve high long-term rewards, whereas the solo competitor B experiences a gradual decline in performance as the coalition’s joint strategy strengthens. Finally, when B and C cooperate against A, the coalition again maintains a sustained reward advantage, and A converges to a lower equilibrium level after an initial transient. Overall, these results demonstrate that cooperation—even when partial—can create persistent performance gaps, with coalition members consistently outperforming the isolated competitor in the long-run equilibrium.

\subsection{Comparison against MOO results}

\begin{figure}[htbp]
    \centering
    \includegraphics[width =0.6\textwidth]{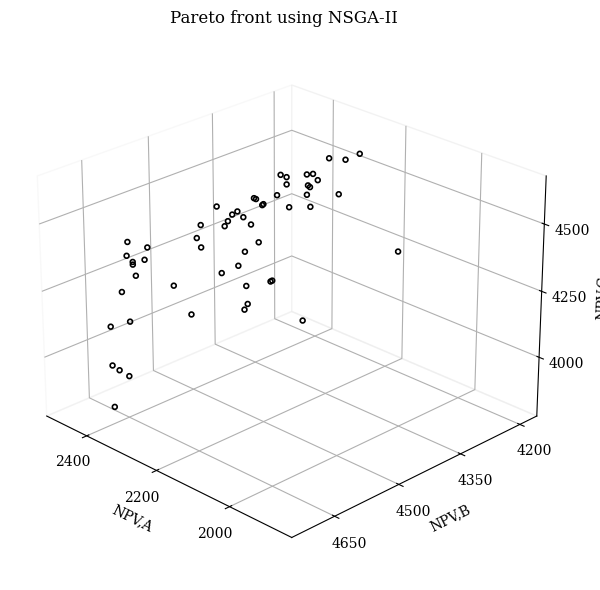}
    \caption{Pareto front obtained using NSGA-II for constrained MOO. The units are in \$ million. } 
    \label{fig: moo_pareto_plots}
\end{figure}

In this subsection, we assess the performance of the proposed constrained Markov game (CMG) framework benchmarked by constrained MOO solutions. The Pareto front produced by NSGA-II is shown in Figure \ref{fig: moo_pareto_plots}, and Table \ref{tab:compare_marl_moo} summarizes the NPVs obtained under different coalition structures and methods. 

As suggested in Table \ref{tab:compare_marl_moo}, the fully cooperative MADDPG setting yields the highest NPVs for all three companies (A: \$2367 M, B: \$4876 M, C: \$4948 M) among different coalition structures. The total NPV reaches its maximum (\$12,191 M) in this setting, showing that end-to-end policy learning can discover highly coordinated strategies that improve overall system efficiency when operators share a common objective. In contrast, fully competitive MADDPG results in substantial decrease of payoffs for each company—most notably for Company C (a \$1385 M decrease compared to the grand coalition setting). The grand competition case also leads to the lowest overall NPV (\$8942 M), illustrating the inefficiency induced when agents optimize selfish objectives without coordination. 

The mixed-coalition MADDPG settings create asymmetric value distributions: when A \& B cooperate against C, both coalition members improve relative to the competitive baseline, especially B (\$4563 M), while C receives similar reward relative to the grand competitive case. Similarly, in A \& C vs. B, the coalition partners (A and C) gain moderately, whereas B is disadvantaged (\$3597 M). In B \& C vs. A, coalition members B and C capture very high NPVs (4750 and 4766), while A receives a low payoff (\$1808 M). Collectively, the mixed-coalition outcomes highlight a consistent distributional effect: forming a coalition increases the coalition members’ NPVs relative to the fully competitive baseline, while the remaining standalone operator is comparatively disadvantaged. In particular, the paired firms in each mixed setting capture most of the value created by coordinated actions, whereas the non-coalition company converges to a lower equilibrium payoff. These results suggest that partial cooperation can yield substantial coalition gains, but it may also amplify payoff asymmetries across operators under decentralized competition.

By comparison, the MOO baseline generates a set of non-dominated trade-off solutions. We note, however, that NSGA-II is a heuristic and may not recover the exact Pareto set in all cases \citep{deb2002fast, yadav2023finding}; here it is therefore used as a practical benchmark. The knee point selection yields relatively high NPVs for all three companies (A: \$2419 M, B: \$4630 M, C: \$4415 M) and a strong total NPV (\$11 464 M). When the selection criterion is adjusted to favor a single company, the favored firm’s NPV increases accordingly, as reflected in the A-favored (\$2475 M), B-favored (\$4709 M), and C-favored (\$4618 M) cases.

Although the cooperative MARL setting optimizes a team reward while MOO explicitly parameterizes multi-objective trade-offs, Table \ref{tab:compare_marl_moo} indicates that the fully cooperative MADDPG outcome is non-dominated within the reported set: it achieves the largest total NPV and attains the highest individual NPVs for Companies B and C (while Company A is slightly lower than the MOO knee-point result). Therefore, within the discrete set of solutions compared in Table \ref{tab:compare_marl_moo}, the cooperative MADDPG result is consistent with a Pareto-efficient outcome. We emphasize that this observation does not constitute a formal proof of global Pareto optimality over the continuous feasible space, which would require verifying dominance against the full Pareto front.

Overall, MADDPG is effective at discovering highly coordinated policies that can maximize coalition-level gains under different coalition structures, whereas MOO provides interpretable, preference-controlled trade-offs that may sacrifice some peak system performance in exchange for more predictable and stakeholder-balanced allocations.




\begin{table}[h!]
\centering
\caption{NPV results (million \$) for different methods.}
\resizebox{\textwidth}{!}{
\begin{tabular}{lcccc}
\hline
\textbf{Methods} & \textbf{Company A} & \textbf{Company B} & \textbf{Company C} & \textbf{Total} \\
\hline
MADDPG, fully collab       & 2367 & 4876 & 4948 & 12191 \\
MADDPG, fully comp         & 1758 & 3621 & 3563 & 8942 \\
MADDPG, {A\&B} v.s. {C}      & 2316 & 4563 & 3624 & 10503 \\
MADDPG, {A\&C} v.s. {B}          & 2300 & 3597 & 4557 & 10454 \\
MADDPG, {B\&C} v.s. {A}        & 1808 & 4750 & 4766 & 11324 \\
\hline
MOO, knee point & 2419 & 4630 & 4415 & 11464 \\
MOO, favoring A & 2475 & 4295 & 4238 & 11008 \\
MOO, favoring B & 2370 & 4709 & 4180 & 11259 \\
MOO, favoring C & 2106 & 4404 & 4618 & 11128 \\
\hline
\end{tabular}
}
\label{tab:compare_marl_moo}
\end{table}

\section{Conclusions}
\label{sec:sec6}
This study develops a constrained Markov game (CMG) formulation and a safe multi-agent deep deterministic policy gradient (MADDPG) implementation for basin-scale geological carbon storage (GCS), where multiple operators interact through shared pressure dynamics and must satisfy safety constraints. To make repeated forward evaluations tractable, we couple the learning framework with an Embed-to-Control (E2C) reduced-order surrogate, enabling efficient training and rollout in a sequential decision-making setting.

Using a three-operator synthetic basin case study, we show that coalition structure strongly shapes both system-level efficiency and the distribution of economic outcomes. Under full cooperation (grand coalition), agents share a team reward and learn coordinated controls that yield the highest total NPV in our experiments (\$12,191 M), whereas full competition produces the lowest total NPV (\$8,942 M), consistent with inefficiencies induced by self-interested responses under interference. In mixed coalition settings, partial cooperation systematically benefits the coalition members relative to the fully competitive baseline while disadvantaging the standalone operator, revealing persistent payoff asymmetries that can emerge when coordination is selective rather than basin-wide. Across all scenarios, the safety-aware learning formulation (cost critics with Lagrangian updates) leads to policies that respect the prescribed pressure limits, supporting the use of CMG/safe-MARL as a practical mechanism for learning decentralized, executable operating rules under regulatory constraints. We further benchmark CMG/safe-MADDPG against constrained multi-objective optimization (CMOO) using NSGA-II. The MOO baseline provides interpretable, preference-controlled trade-offs and yields comparatively balanced allocations, while the cooperative MADDPG outcome achieves the highest total NPV and is non-dominated within the discrete set of solutions reported. 

Our future endeavor will focus on addressing some limitations of current study. First, while pressure is a widely used and illustrative safety constraint, the proposed CMG framework is not restricted to pressure and can incorporate additional operational or regulatory constraints (e.g., plume extent/AoR, leakage-risk indicators) through appropriate cost definitions.Second, the case study is designed to test the decision-making paradigm; scaling to field settings will require robust surrogates (or other efficient forward models), proper adjustment of the code algorithm, and so on. Finally, although the cooperative MADDPG solution is Pareto-efficient within the compared discrete set, establishing global Pareto optimality over the continuous feasible space would require explicit verification against the full Pareto front. Developing MARL variants with stronger Pareto-efficiency guarantees and/or integrating preference articulation into learning are promising directions.



\section*{Acknowledgments}
\label{sec:acknowledgement}
We acknowledge the Gulf Coast Carbon Center at UT Austin and its sponsors for funding this work. We thank Sean Avitt for providing the Wilcox Group geomodel in Figure \ref{fig: marl_framework}. Thanks also go to Computer Modeling Group Ltd. for providing the simulation software. Finally, we appreciate the helpful discussions with Dr. Per Pettersson and Dr. Sarah Gasda regarding game-theoretic applications in CCS projects. 

\section*{Data Availability}
\noindent
The code will be made available at
\href{https://github.com/jungangc/CCS_MARL}{https://github.com/jungangc/CCS\_MARL} once published.

\section*{Declaration of generative AI and AI-assisted technologies in the manuscript preparation process}
\noindent
During the preparation of this manuscript, the authors used ChatGPT 5.1 to assist with language editing. All content generated with this tool was subsequently reviewed, revised, and approved by the authors, who take full responsibility for the final published work.

\bibliographystyle{elsarticle-harv} 
\bibliography{Ref}

\section*{Appendix}

\begin{figure}[htbp]
    \centering
    \includegraphics[width=0.6\textwidth]{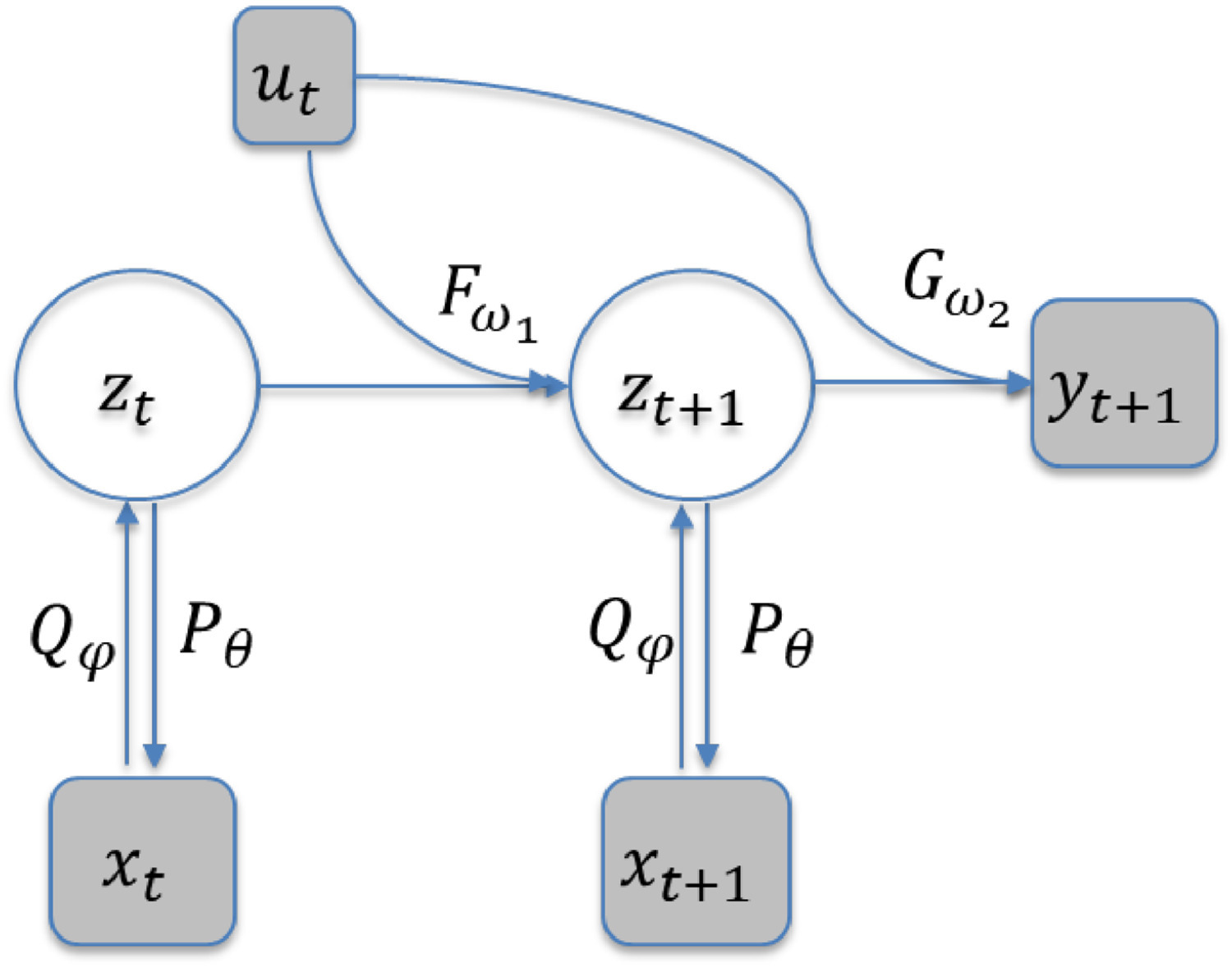}
    \caption{Graphical representation of E2C model, adapted from \citep{chen2024optimization}}
    \label{fig: e2co_plot}
\end{figure}

Figure \ref{fig: e2co_plot} illustrates the graphical structure of the E2C model, where the gray boxes denote the training tuple 
$\{x_t,\, u_t,\, x_{t+1},\, y_{t+1}\}$. The encoder network $Q_{\varphi}$ maps inputs into a latent representation, 
while the decoder network $P_{\theta}$ reconstructs the corresponding outputs. The transition model, expressed as 
$F_{\omega_1}$, predicts the next latent state $z_{t+1}$ from the current latent state $z_t$ and the control input $u_t$. 
In practice, this transition function is implemented using two neural networks that approximate the matrices $A$ and $B$, 
thereby enabling the use of a locally linear dynamical equation. Likewise, the output network $G_{\omega_2}$ is designed 
to approximate the matrices $C$ and $D$. For further implementation and validation details, readers are referred to \citep{chen2024optimization}.

\begin{figure}[htbp]
    \centering
    \includegraphics[width=0.6\textwidth]{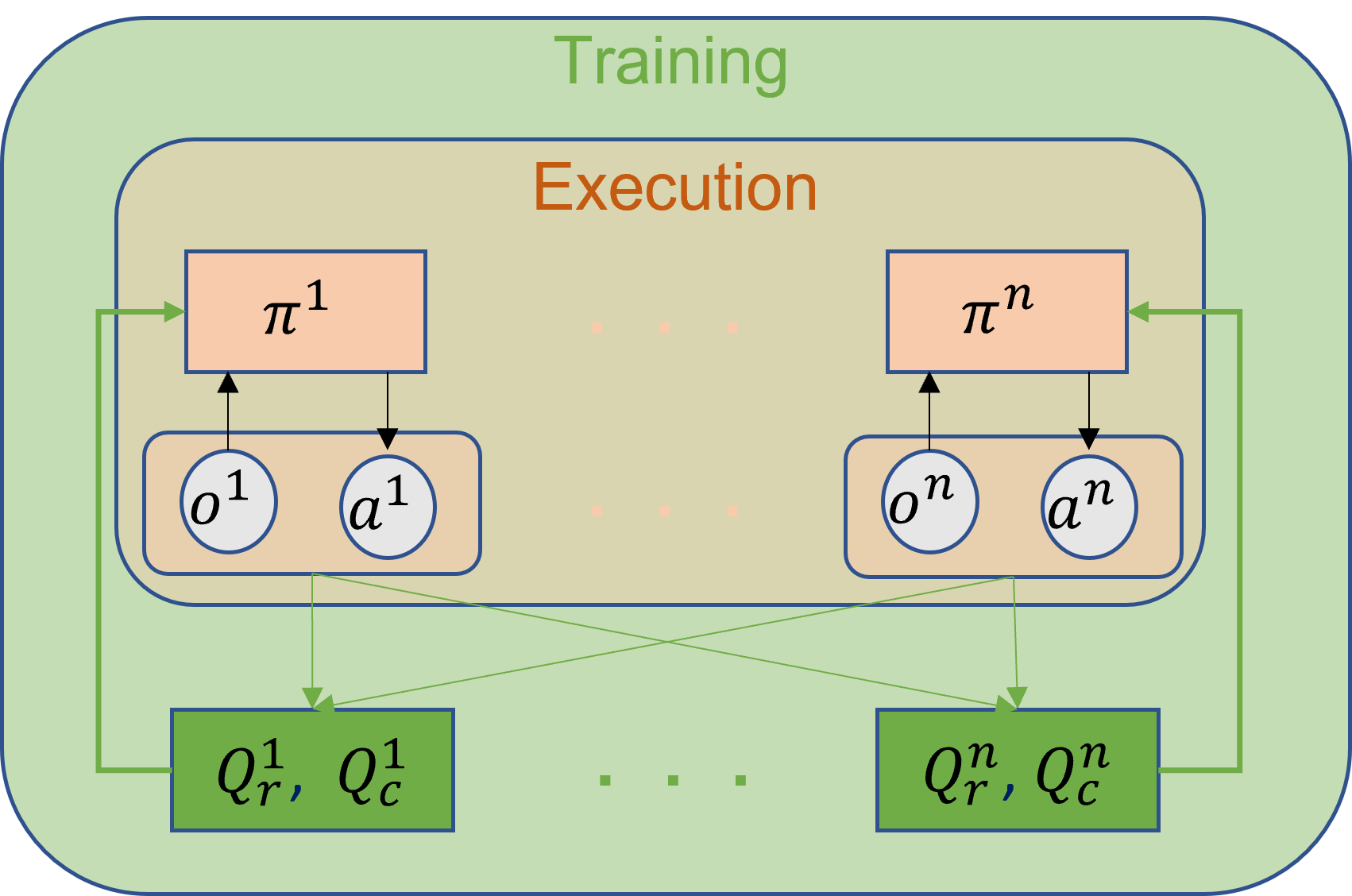}
    \caption{Safe MADDPG with centralized training and decentralized execution (CTDE) \citep{lowe2017multi}. Inside the \emph{Execution} box, each agent $i$ selects $a_i=\pi_i(o_i)$ from its local observation. 
    During \emph{Training} (outer box), a reward critic $Q^r_{i}$ and a cost critic $Q^c_{i}$ for each agent condition on the joint tuple $(o_{1:n},a_{1:n})$ and backpropagate gradients to the actors. 
    Safety is enforced via a Lagrangian penalty $-\lambda_i Q^c_{i}$, with multipliers $\lambda_i$ updated to penalize constraint violations; at evaluation time only the policies run, while critics (and $\lambda_i$ updates) are not required.}
    \label{fig: maddpg_scheme}
\end{figure}

\begin{figure}[htbp]
    \centering
    \includegraphics[width =\textwidth]{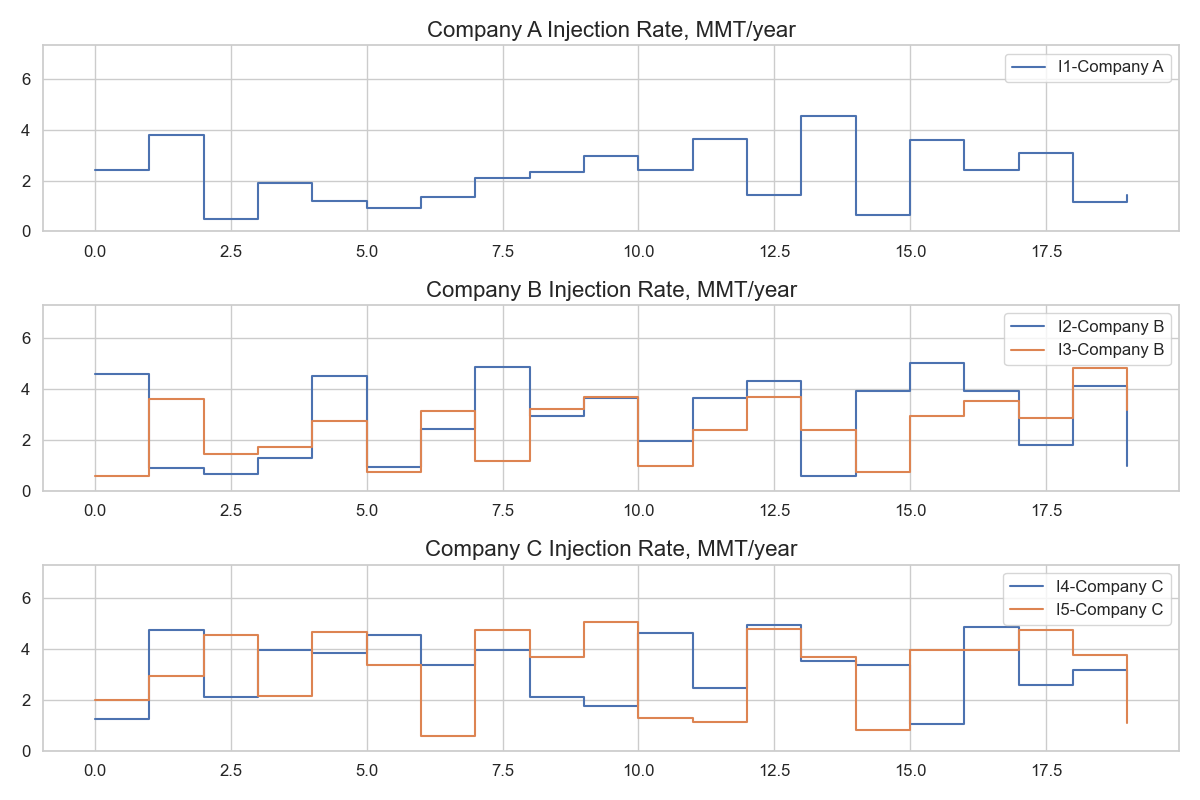}
    \caption{Random well control strategies employed by all companies. The x axis indicates the injection time in years. } 
    \label{fig: base_actions}
\end{figure}

\begin{figure}[htbp]
    \centering
    \includegraphics[width =1.1\textwidth]{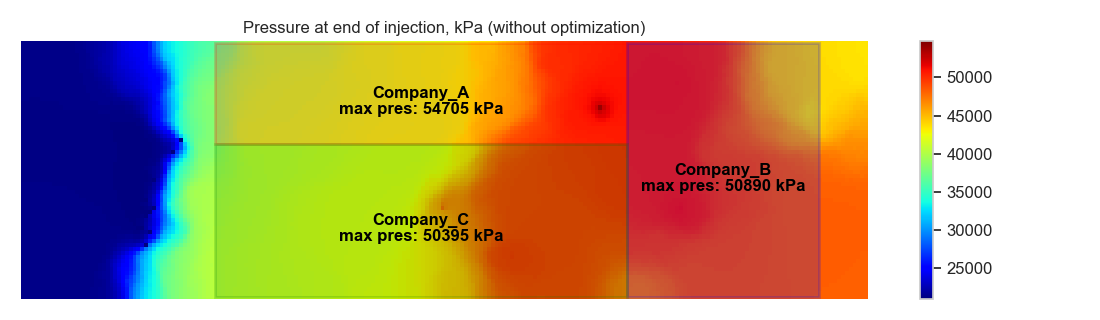}
    \caption{Pressure map at the end of injection if using arbitrary well control strategies in figure \ref{fig: base_actions}.} 
    \label{fig: base_pressure}
\end{figure}

Figure \ref{fig: maddpg_scheme} shows the centralized training decentralized execution scheme in MADDPG. Figure \ref{fig: base_actions} displays the well control strategies of all 3 companies over 20 years, while figure \ref{fig: base_pressure} presents the corresponding reservoir pressure distribution at the end of injection.

\begin{algorithm}[t]
\caption{Constrained MADDPG for multi--stakeholder \texorpdfstring{CO$_2$}{CO2} management}
\label{alg:cmaddpg}
\footnotesize
\setlength{\abovedisplayskip}{2pt}
\setlength{\belowdisplayskip}{2pt}
\setlength{\abovedisplayshortskip}{2pt}
\setlength{\belowdisplayshortskip}{2pt}

\begin{algorithmic}[1]
\Require episodes $M$, horizon $T$, discount $\gamma$, soft update rate $\tau$, replay buffer $\mathcal D$, exploration noise $\mathcal N$, safety threshold $\bar c$, dual stepsize $\eta_\lambda$, max multiplier $\lambda_{\max}$, actor $\mu_i$ with parameters $\theta_i$, reward critic $Q^{r}_i$ with params $\phi^{r}_i$, penalty critic $Q^{c}_i$ with params $\phi^{c}_i$
\State \textbf{Initialize:} target networks $\theta'_i \leftarrow \theta_i$, $\phi^{r\prime}_i \leftarrow \phi^{r}_i$, $\phi^{c\prime}_i \leftarrow \phi^{c}_i$; $\lambda \ge 0$
\For{$\text{episode}=1\ \text{to}\ M$}
  \State Reset environment; get initial observations $o_1,\ldots,o_N$
  \For{$t=1\ \text{to}\ T$}
    \State \textbf{(Act)} For each agent $i$, select action
      $a_i \gets \mu_i(o_i) + \mathcal N_t$
    \State Execute joint action $\mathbf a=(a_1,\ldots,a_N)$; observe reward vector $\mathbf r$, penalty vector $\mathbf c$ and next observations $\mathbf o'$ according to equations \ref{eqn: reward}, \ref{formula: penalty_well} and \ref{eqn: reducedstatespace}, . 
    \State Store transition $(\mathbf o,\mathbf a,\mathbf r,\mathbf c,\mathbf o')$ in buffer $\mathcal D$; set $\mathbf o \gets \mathbf o'$
    \Statex
    \State \textbf{(Learn)} Sample a minibatch $\{(\mathbf o^j,\mathbf a^j,\mathbf r^j,\mathbf c^j,\mathbf o^{\prime j})\}_{j=1}^S \sim \mathcal D$
    \For{$i=1$ to $N$} \Comment{update per agent}
      \State $y^{r,j}_i \gets r_i^j + \gamma \, Q^{r\prime}_i(\mathbf o^{\prime j}, a^{\prime j}_1,\ldots,a^{\prime j}_N)\Big|_{a^{\prime j}_k \gets \mu'_k(o^{\prime j}_k)}$
      \State $y^{c,j}_i \gets c_i^j + \gamma \, Q^{c\prime}_i(\mathbf o^{\prime j}, a^{\prime j}_1,\ldots,a^{\prime j}_N)\Big|_{a^{\prime j}_k \gets \mu'_k(o^{\prime j}_k)}$
      \State Update critics by minimizing 
        \[
        \mathcal L^{r}_i(\phi^{r}_i)=\frac{1}{S}\sum_{j=1}^S\!\Big(y^{r,j}_i - Q^{r}_i(\mathbf o^{j}, \mathbf a^{j})\Big)^2,\quad
        \mathcal L^{c}_i(\phi^{c}_i)=\frac{1}{S}\sum_{j=1}^S\!\Big(y^{c,j}_i - Q^{c}_i(\mathbf o^{j}, \mathbf a^{j})\Big)^2
        \]
      \State Update actors with sampled policy gradient
        \[
        J_i(\theta_i)\;=\; \frac{1}{S}\sum_{j=1}^S
        \Big[-\,Q^{r}_i(\mathbf o^{j}, a^{j}_1,\ldots,a^{j}_N)
        + \lambda\, Q^{c}_i(\mathbf o^{j}, a^{j}_1,\ldots,a^{j}_N)\Big]\Big|_{a^{j}_k \gets \mu'_k(o^{j}_k)}
        \]
     \State dual update for Lagrange multiplier:
      \[
      \lambda \leftarrow max\Big(0, \lambda + \eta_\lambda\,\big(\frac{1}{NS}\!\sum_{i=1}^N\!\sum_{j=1}^S\big( Q^{c}_i- \bar c\big)\Big)
      \]
    \EndFor

    \State \textbf{Target soft updates} (for each $i$):
      \[
      \theta'_i \leftarrow \tau\,\theta_i + (1-\tau)\,\theta'_i,\quad
      \phi^{r\prime}_i \leftarrow \tau\,\phi^{r}_i + (1-\tau)\,\phi^{r\prime}_i,\quad
      \phi^{c\prime}_i \leftarrow \tau\,\phi^{c}_i + (1-\tau)\,\phi^{c\prime}_i
      \]
  \EndFor
\EndFor
\end{algorithmic}
\end{algorithm}

\end{document}